\documentclass[10pt,twocolumn,letterpaper]{article}

\usepackage[pagenumbers]{cvpr} %

\usepackage[dvipsnames]{xcolor}
\usepackage{colortbl}

\definecolor{placeholder}{rgb}{0.6,0.8,0.95}

\definecolor{amber}{rgb}{1.0, 0.75, 0.0}

\definecolor{visible-blue}{rgb}{0.286, 0.525, 0.910}

\definecolor{tabfirst}{rgb}{1, 0.7, 0.7} 
\definecolor{tabsecond}{rgb}{1, 0.85, 0.7} 
\definecolor{tabthird}{rgb}{1, 1, 0.7} 

\definecolor{cvprblue}{rgb}{0.21,0.49,0.74}
\usepackage[pagebackref,breaklinks,colorlinks,citecolor=cvprblue]{hyperref}
\usepackage{pifont}
\usepackage{color}
\usepackage{xcolor}
\usepackage{nicefrac}       %
\usepackage{multirow}
\usepackage{multicol}
\usepackage{dsfont}

\usepackage{colortbl}
\usepackage[percent]{overpic}

\usepackage{bbm}
\usepackage{algorithm}
\usepackage{algpseudocode}

\definecolor{3dgsblue}{RGB}{0,176,240}
\definecolor{mipyellow}{RGB}{255,192,0}
\definecolor{oursred}{RGB}{239,148,158}

\def\paperID{****} 
\def\confName{CVPR}
\def\confYear{2024}

\title{Spectral-GS: Taming 3D Gaussian Splatting with Spectral Entropy}

\author{Letian Huang \quad Jie Guo\protect\footnotemark[2] \quad Jialin Dan \quad Ruoyu Fu \quad Shujie Wang \quad Yuanqi Li \quad Yanwen Guo \\
Nanjing University\\
{\tt\small \{lthuang, danjialin, wangshujie\}@smail.nju.edu.cn},\quad
{\tt\small \{guojie, fry, yuanqili, ywguo\}@nju.edu.cn}
}
\begin{document}

\twocolumn[{%
	\renewcommand
	\twocolumn[1][]{#1}%
	\maketitle
	\begin{center}
		\centering
		\vspace{-15pt}
            \includegraphics[width=\textwidth]{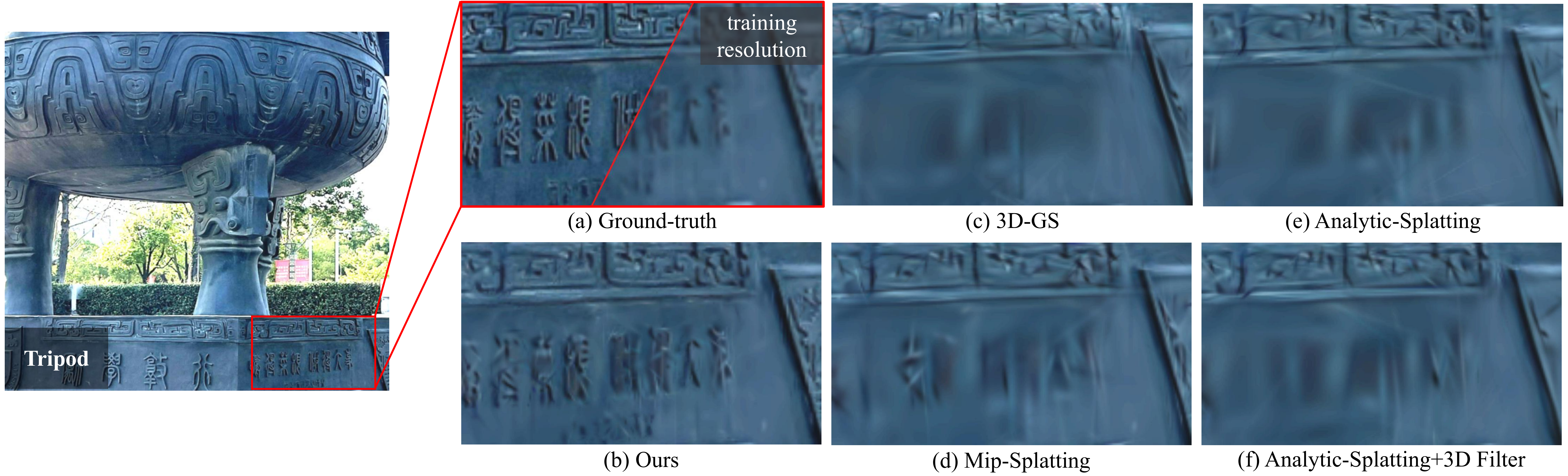}
		\captionof{figure}{\small
Despite its high efficiency in 3D reconstruction, 3D Gaussian Splatting (3D-GS)~\cite{kerbl20233d} suffers from needle-like artifacts \textbf{(c)} due to undersampling or view inconsistency. Recent works like Mip-Splatting~\cite{mip_gs} and Analytic-Splatting~\cite{liang2024analytic} try to eliminate these artifacts. Unfortunately, they still produce needles at high-frequency regions when zooming in, and will  also cause over-blurriness \textbf{(d)(e)} since they lack shape awareness of 3D Gaussians. With spectral analysis of the variance matrix, we propose \textbf{Spectral-GS}, which imposes shape constraints on the 3D Gaussians and thus effectively addresses the above issues, generating high-quality photorealistic rendering \textbf{(b)}.
		} 
		\label{fig:teaser}
	\end{center}
}]
\maketitle

\begingroup
\renewcommand{\thefootnote}{\fnsymbol{footnote}}
\footnotetext[2]{Corresponding author.}
\endgroup
\begin{abstract}
\vspace{-0.3cm}
Recently, 3D Gaussian Splatting (3D-GS) has achieved impressive results in novel view synthesis, demonstrating high fidelity and efficiency. However, it easily exhibits needle-like artifacts, especially when increasing the sampling rate. Mip-Splatting tries to remove these artifacts with a 3D smoothing filter for frequency constraints and a 2D Mip filter for approximated supersampling. Unfortunately, it tends to produce over-blurred results, and sometimes needle-like Gaussians still persist. Our spectral analysis of the covariance matrix during optimization and densification reveals that current 3D-GS lacks shape awareness, relying instead on spectral radius and view positional gradients to determine splitting. As a result, needle-like Gaussians with small positional gradients and low spectral entropy fail to split and overfit high-frequency details. Furthermore, both the filters used in 3D-GS and Mip-Splatting reduce 
 the spectral entropy and increase the condition number during zooming in to synthesize novel view, causing view inconsistencies and more pronounced artifacts. Our Spectral-GS, based on spectral analysis, introduces 3D shape-aware splitting and 2D view-consistent filtering strategies, effectively addressing these issues, enhancing 3D-GS's capability to represent high-frequency details without noticeable artifacts, and achieving high-quality photorealistic rendering.
\end{abstract}    
\section{Introduction}
\label{sec:intro}

Reconstructing 3D scenes from 2D images and synthesizing novel views has been a critical task in computer vision and graphics. As the demand for real-time and photo-realistic rendering continues to rise, 3D Gaussian Splatting (3D-GS)~\cite{kerbl20233d} has emerged as an efficient representation that can achieve high-speed rendering on the GPU. Unlike traditional implicit scene representations using MLPs~\cite{mildenhall2020nerf, nerfpp, barron2021mip, barron2022mip}, 3D-GS adopts an explicit approach based on Gaussian functions. This bypasses the need for dense point sampling in volumetric rendering, thereby enabling real-time performance. However, 3D-GS~\cite{kerbl20233d} tends to optimize toward degraded needle-like Gaussians, resulting in unacceptable needle-like artifacts. Mip-Splatting~\cite{mip_gs} and Analytic-Splatting~\cite{liang2024analytic} attempt to address these issues by employing filtering or analytic integration to mitigate aliasing. Unfortunately, when representing high-frequency textures, these methods~\cite{mip_gs, liang2024analytic} often lead to over-blurriness or still produce needle-like artifacts. 

Needle-like Gaussians correspond to 3D Gaussians with low spectral entropy and high condition number. Existing 3D-GS~\cite{kerbl20233d} and its variants (including Mip-Splatting~\cite{mip_gs}) do not impose any restriction on the Gaussian's shape. The splitting strategies in 3D-GS lean towards generating degenerated elongated Gaussians, due to the lack of shape-awareness, relying instead on spectral radius and view positional gradients to guide the splitting. However, needle-like Gaussians with small positional gradients are hard to split. Even splitting, the condition number of the Gaussian remains ill-conditioned providing little help to alleviate needle-like artifacts. Worse still, both the EWA filter~\cite{zwicker2002ewa} in 3D-GS~\cite{kerbl20233d} and the 2D Mip filter in Mip-Splatting~\cite{mip_gs} reduce the spectral entropy and increase the condition number during zooming in to synthesize novel view. Due to the view-inconsistency in filtering, needle-like artifacts become more pronounced when zooming in or when the camera moves closer to the object.

Based on the above observations, we introduce \textit{spectral analysis} to 3D Gaussian reconstruction. Specifically, we propose \textit{3D shape-aware splitting} and
\textit{2D view-consistent filtering}, respectively, to address loss sensitivity and shape unawareness in densification, as well as view inconsistency in filtering. The splitting condition of our 3D shape-aware splitting is based on the spectral entropy of 3D Gaussians and our method ensures that the condition number after splitting reduces. The proposed 2D view-consistent filtering, combines a convolution that approximates supersampling with a view-adaptive Gaussian blur that approximates interpolation to maintain the spectral entropy consistency. Our method Spectral-GS effectively enhances 3D-GS's capability to represent high-frequency details, mitigates needle-like artifacts, and achieves high-quality photorealistic rendering, as illustrated in Figure~\ref{fig:teaser}. Furthermore, our method is easily implemented, requiring only few changes to the original framework. In summary, we make the following contributions:

\begin{itemize}
    \item We employ \textbf{spectral analysis} to examine 3D-GS, revealing issues such as loss sensitivity and shape unawareness in densification, as well as view inconsistency in filtering.
    \item We propose \textbf{3D shape-aware splitting} to regularize needle-like Gaussians, enhancing the high-frequency detail representation for 3D-GS and mitigating needle-like artifacts.
    \item We propose \textbf{2D view-consistent filtering} to resolve needle-like artifacts caused by view-inconsistency.
\end{itemize}
\section{Related Work}

Novel view synthesis is a longstanding challenge in computer vision and graphics.  From traditional techniques ~\cite{schonberger2016structure, DBLP:conf/siggraph/GortlerGSC96, DBLP:conf/siggraph/LevoyH96, DBLP:conf/siggraph/BuehlerBMGC01,Sfm_init,DBLP:conf/iccv/GoeseleSCHS07, DBLP:journals/tog/ChaurasiaDSD13}, to neural network-based scene representations~\cite{jiang2020,deepsdf,genova2020,occupancynet,diffvolumetric,srn,DBLP:journals/tog/KopanasLRJD22}, various approaches have struggled to address the problem of synthesizing a new view from captured images. 



\subsection{Neural Radiance Fields} 
The Neural Radiance Field (NeRF)~\cite{mildenhall2020nerf} stands out as a successful neural rendering method based on MLPs, primarily owing to its encoding of position and direction. This encoding allows for effective reconstruction of high-frequency information in scenes. Notable improvements on this encoding have been made by MipNeRF~\cite{barron2021mip}, NeRF-W~\cite{nerfw}, FreeNeRF~\cite{freenerf}, and Instant NGP~\cite{muller2022instant}. These enhancements enable the handling of multi-resolution image inputs, multi-illumination with occlusion image inputs, sparse-view inputs, and achieve nearly real-time rendering capabilities, respectively. 
Barron \etal introduced MipNeRF360~\cite{barron2022mip} as an extension of MipNeRF~\cite{barron2021mip} to address the issue of generating low-quality renderings for unbounded scenes in NeRF. Importantly, these methods thoroughly exploit the intrinsic capabilities of NeRF as an implicit scene representation without introducing additional model priors. However, due to the implicit representation of scenes and the dense sampling of points along rays, they still face challenges in achieving real-time performance. 

\subsection{3D Gaussian Splatting} 3D Gaussian Splatting (3D-GS)~\cite{kerbl20233d} has demonstrated notable advancements in rendering performance by departing from MLPs and ray sampling, opting instead for anisotropic Gaussians and projection. This paradigm shift has attracted considerable attention within the industry, leading to various studies upon this method~\cite{chen2024survey, gs_slam, gs_scene_edit, gs_scene_seg, dynamic_gs}.


Recent endeavors aim to enhance the robustness in sparse-view scenarios~\cite{xiong2023sparsegs,yang2024gaussianobject}, performance, storage efficiency~\cite{niedermayr2023compressed} and mesh reconstruction~\cite{sugar} of 3D-GS. However, none of these methods specifically address the needle-like artifacts that are absent in NeRF but present in 3D-GS. Some methods~\cite{huang2024erroranalysis3dgaussian, mip_gs} are exploring the potential of 3D-GS. Nevertheless, the former~\cite{huang2024erroranalysis3dgaussian} proposes the optimal projection to address artifacts caused by the increased projection error with a larger field of view. Mip-Splatting~\cite{mip_gs} proposes filtering to address aliasing caused by sampling. Analytic-Splatting~\cite{liang2024analytic} replaces Mip-Splatting's 2D Mip filter with a closed-form expression for the Gaussian integral within a pixel. These methods~\cite{mip_gs, liang2024analytic} partially alleviate the needle-like artifacts. However, they still rely on the conventional density control solely based on the spectral radius and loss gradients without shape-awareness, leading to loss of high-frequency details.


\begin{figure}[t]

\includegraphics[width=1\linewidth]{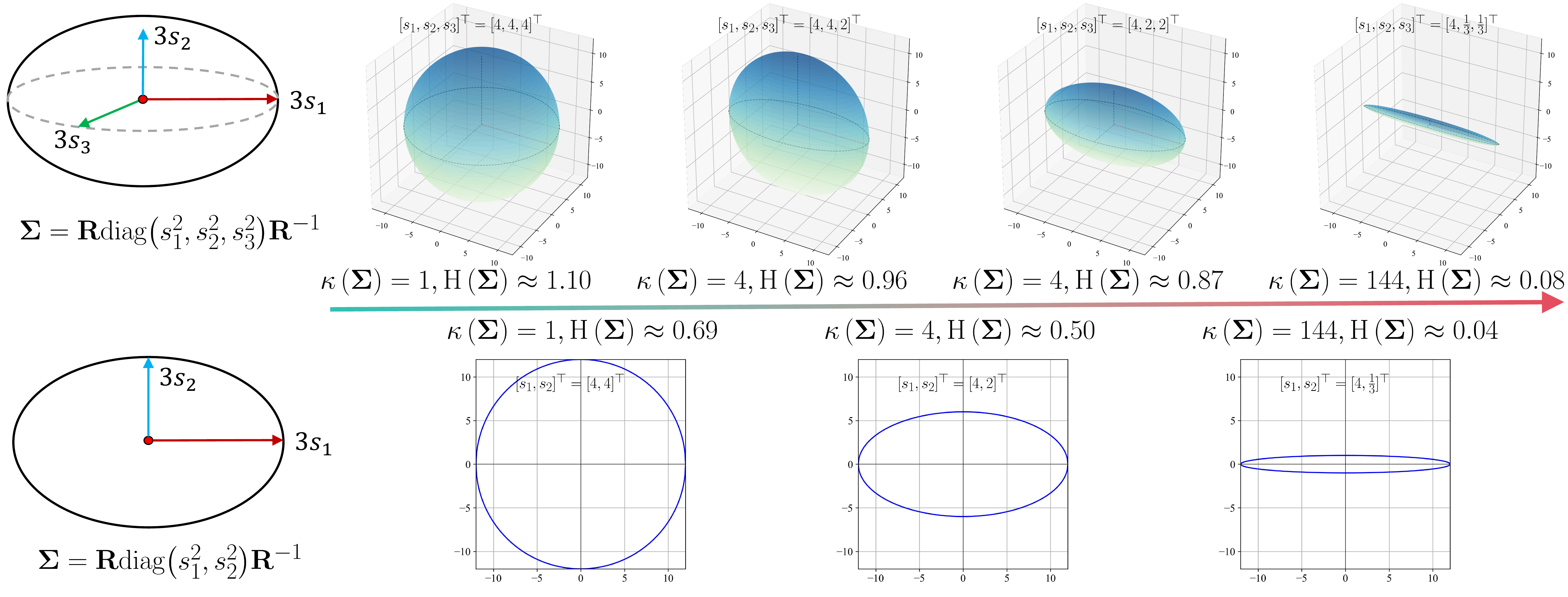}
\caption{Visualization of Gaussians with the same spectral radius but different shapes. The spectrum of the 3D Gaussian is characterized by $s_1, s_2, s_3$ (top row), while the 2D Gaussian is characterized by $s_1, s_2$ (bottom row). From left to right, as 
 the spectral entropy decreases and the condition number increases, the Gaussians transition from isotropic to anisotropic.} 
\label{fig:spectrum}
\end{figure}

\section{Preliminaries}


\label{subsec:3dgs}


\vspace{0.1cm}\noindent{\bf Representation and Projection:}
3D Gaussian Splatting~\cite{kerbl20233d} constructs a scene representation using volumetric primitives $\mathcal{G}^{3D}\left(\cdot\right)$, each characterized by position (also known as the mean) $\boldsymbol{\mu}$, a covariance matrix $\boldsymbol{\Sigma}$ (decomposed into scale $\mathbf{S}\in \mathbb{R}^{3\times3}$ and rotation $\mathbf{R} \in \text{SO}\left(3\right)$, \ie,
$\boldsymbol{\Sigma}=\mathbf{R}\mathbf{S}\mathbf{S}^{\top}\mathbf{R}^{\top}$), opacity $o$, and spherical harmonics coefficients $\text{SH}\left(\cdot\right)$. The 3D primitives are projected to 2D image space through a Jacobian matrix $\mathbf{J}\in \mathbb{R}^{2\times3}$ for the local affine approximation after being transformed to the camera space via a viewing transformation
matrix $\mathbf{W}$. Then 2D Gaussians $\mathcal{G}^{2D}\left(\cdot\right)$, each characterized by the position $\boldsymbol{\mu}_{\text{proj}}$ and a covariance matrix $\boldsymbol{\Sigma}_{\text{proj}}=\mathbf{J}\mathbf{W}\boldsymbol{\Sigma}\mathbf{W}^{\top}\mathbf{J}^{\top}$, are rasterized using $\alpha$-blending. 

\vspace{0.1cm}\noindent{\bf Optimization and Densification:} 3D-GS~\cite{kerbl20233d} employs a loss function that combines $\mathcal{L}_1$ loss with a D-SSIM term:
\begin{equation}
    \mathcal{L} = (1-\lambda_1)\mathcal{L}_1 + \lambda_1 \mathcal{L}_{\text{D-SSIM}}.  
\end{equation} For this loss can only optimize the parameters of Gaussian primitives but cannot change the number of primitives, phenomena such as “over-reconstruction” (regions where Gaussians cover large areas in the scene) and “under-reconstruction” (regions with missing geometric features) can occur. To address this, 3D-GS introduces an adaptive Gaussian densification scheme. For Gaussians with large view-space positional gradients $\nabla_{\boldsymbol{\mu}_{\text{proj}}}\mathcal{L}$, the scheme chooses between clone and split strategies based on the scale of Gaussians. After splitting, the shape remains unchanged, with the scale being $\frac{1}{k}$ ($k=1.6$ in all experiments) of the original, \ie, $\boldsymbol{\Sigma}_{\text{split}}=\frac{1}{k^2}\boldsymbol{\Sigma} $.


\vspace{0.1cm}\noindent{\bf Filtering and Mip-Splatting~\cite{mip_gs}:} To prevent projected Gaussians from becoming too small to cover an entire pixel, 3D-GS uses an EWA filter~\cite{zwicker2002ewa} for training stability:
\begin{gather}
\mathcal{G}_{k}^{2D}\left(\mathbf{u}\right)_{\text{EWA}}=oe^{-\frac{1}{2}\left(\mathbf{u}-\boldsymbol{\mu}_{\text{proj}}\right)^{\top}
{\left(\boldsymbol{\Sigma}_{\text{proj}}+s\mathbf{I}\right)}^{-1}
\left(\mathbf{u}-\boldsymbol{\mu}_{\text{proj}}\right)}
\label{eq:ewa_filter}
\end{gather}
where $\mathbf{u}$ is the pixel coordinate, $\mathbf{I}$ is a 2D identity matrix, $s$ is a scalar hyperparameter to control the size of the filter, and $\mathcal{G}_{k}^{2D}\left(\cdot\right)_{\text{EWA}}$  is the EWA filtered Gaussian.

To limit the maximal frequency for the 3D representation, Mip-Splatting~\cite{mip_gs} applies a 3D smoothing filter $\mathcal{G}_{\text{low}}$ to the 3D Gaussians $\mathcal{G}^{3D}$, ensuring that the regularized Gaussians cover at least one pixel in all training views:
\begin{gather}
\begin{split}
&\mathcal{G}_{k}^{3D}\left(\mathbf{x}\right)_{\text{reg}}=\left(\mathcal{G}^{3D}\otimes\mathcal{G}_{\text{low}}\right)\left(\mathbf{x}\right)\\
&=o\sqrt{\frac{\left|\boldsymbol{\Sigma}\right|}{\left|\boldsymbol{\Sigma}+\frac{s}{\hat{\nu}_k}\cdot\mathbf{I}\right| }}e^{-\frac{1}{2}\left(\mathbf{x}-\boldsymbol{\mu}\right)^{\top}
{\left(\boldsymbol{\Sigma}+\frac{s}{\hat{\nu}_k}\cdot\mathbf{I}\right)}^{-1}
\left(\mathbf{x}-\boldsymbol{\mu}\right)}.
\end{split}
\end{gather} Here, the scale $\frac{s}{\hat{\nu}}$ of the 3D filters for each primitive are different as they depend on the training views in which they are visible. In addition, the box filter of a function is equivalent to the integral over the corresponding region, while the convolution of Gaussians remains a Gaussian. Hense, Mip-Splatting~\cite{mip_gs} proposes a 2D Mip filter to approximate the integral or super-sampling within a pixel:
\begin{gather}
\mathcal{G}_{k}^{2D}\left(\mathbf{u}\right)_{\text{Mip}}=o\sqrt{\frac{\left|\boldsymbol{\Sigma}_{\text{proj}}\right|}{\left|\boldsymbol{\Sigma}_{\text{proj}}+s\mathbf{I}\right| }}e^{-\frac{1}{2}\left(\mathbf{u}-\boldsymbol{\mu}_{\text{proj}}\right)^{\top}
{\left(\boldsymbol{\Sigma}_{\text{proj}}+s\mathbf{I}\right)}^{-1}
\left(\mathbf{u}-\boldsymbol{\mu}_{\text{proj}}\right)}
\label{eq:mip_filter}
\end{gather} where $\mathcal{G}_{k}^{2D}\left(\cdot\right)_{\text{Mip}}$ is the 2D Mip filtered Gaussian. To compute the Gaussian integral within a pixel more analytically, Analytic-Splatting~\cite{liang2024analytic} derives the cumulative distribution function (CDF) of the Gaussians, replacing the 2D Mip filter used to approximate a box filter.

\begin{figure}[t]

\includegraphics[width=1\linewidth]{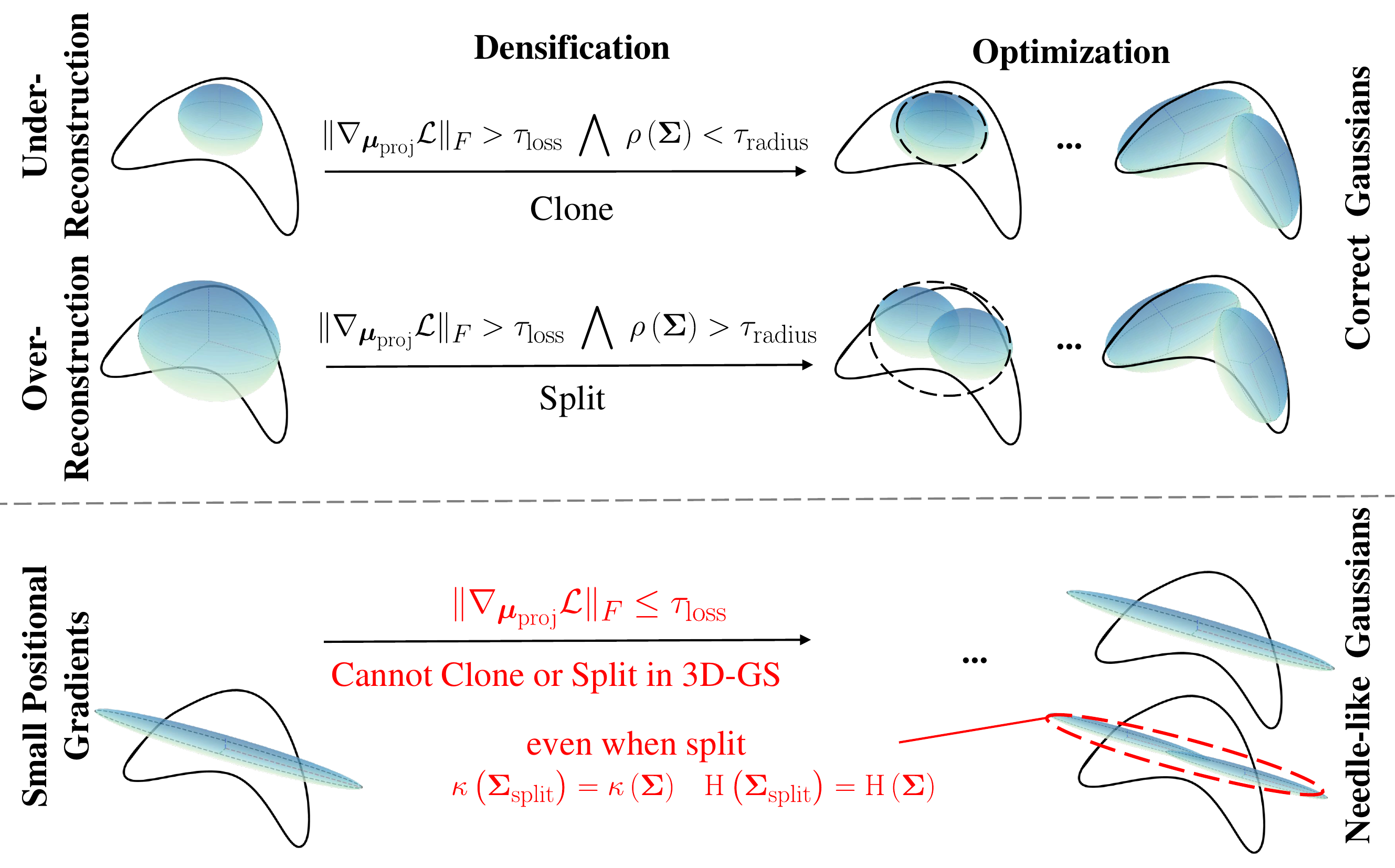}
\caption{Illustrations of the optimization and densification of Gaussians in 3D-GS~\cite{kerbl20233d}. \textit{Correct Gaussians: } When view-positional gradients 
 $\nabla_{\boldsymbol{\mu}_{\text{proj}}}\mathcal{L}$ exceed a certain threshold $\tau_{\text{loss}}$, 3D-GS decides to clone or split based on the Gaussian's spectral radius $\rho\left(\boldsymbol{\Sigma}\right)$. \textit{Needle-like Gaussians: } However, 3D-GS does not split or clone Gaussians with low spectral entropy but small gradients.} 
\label{fig:issue_split}
\end{figure}

\section{Spectral Analysis of 3D Gaussian Splatting}
\label{sec:spectral_analysis_3dgs}



\begin{figure}[t]

\includegraphics[width=1\linewidth]{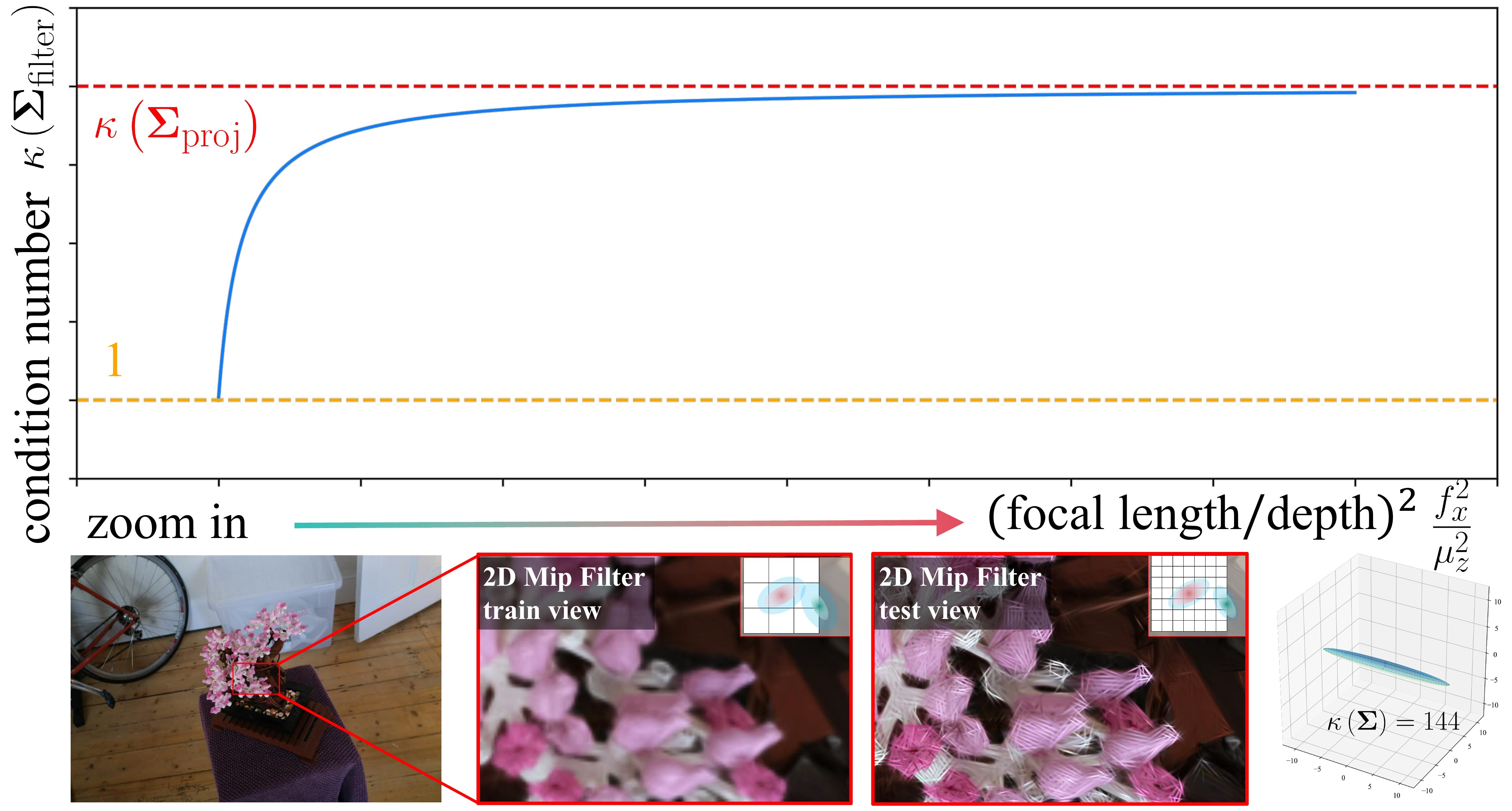}
\caption{Illustrations of the condition number variation curve and rendering results when zooming in. We fix the condition number of 3D Gaussians $\kappa\left(\boldsymbol{\Sigma}\right)=144$ during training. Due to view-inconsistency in filtering, the train view still produces satisfactory rendering results (2D Mip Filter train view), but the test view with higher $\frac{f_x^2}{\mu_z^2}$ shows needle-like artifacts (2D Mip Filter test view).} 
\label{fig:zoomin}
\end{figure}

In this section, we first perform spectral analysis to 3D-GS and show loss-sensitivity and shape-unwareness in densification, and view-inconsistency in filtering.

\vspace{0.1cm}\noindent{\bf Spectra of Gaussians:} The covariance matrix $\boldsymbol{\Sigma}$ of a 3D/2D Gaussian is analogous to describing the configuration of an ellipsoid/ellipse. For example, the covariance matrix of 3D Gaussian can be eigendecomposed as follows:
\begin{equation}
\boldsymbol{\Sigma}=\mathbf{R}\mathbf{S}\mathbf{S}^{\top}\mathbf{R}^{\top}
=\mathbf{R}\left(\mathbf{S}\mathbf{S}^{\top}\right)\mathbf{R}^{-1}=
\mathbf{R}\text{diag}\begin{matrix}\left(s_1^2,s_2^2,s_3^2\right)\end{matrix}\mathbf{R}^{-1}
\end{equation} where the rotation matrix $\mathbf{R}$ is an orthogonal matrix ($\mathbf{R}^{\top}=\mathbf{R}^{-1}$), $3s_1, 3s_2, 3s_3$ represent the lengths of the ellipsoid's three axes ($3\sigma$ rule)~\cite{kerbl20233d}, and $s_1^2, s_2^2, s_3^2$ are the eigenvalues (spectrum) of $\boldsymbol{\Sigma}$. The spectral radius of the covariance matrix $\boldsymbol{\Sigma}$ can be derived as:
\begin{equation}
\rho\left(\boldsymbol{\Sigma}\right)=\max\left(s_1^2,s_2^2,s_3^2\right)
\end{equation} and can be used to measure the scale of the Gaussian. Additionally, the condition number and spectral entropy are respectively given by
\begin{gather}
\kappa\left(\boldsymbol{\Sigma}\right)=\frac{\rho\left(\boldsymbol{\Sigma}\right)}{\rho_{\text{min}}\left(\boldsymbol{\Sigma}\right)}=\frac{\max\left(s_1^2,s_2^2,s_3^2\right)}{\min\left(s_1^2,s_2^2,s_3^2\right)},\\
\text{H}\left(\boldsymbol{\Sigma}\right)=\text{tr}\left(-\frac{\boldsymbol{\Sigma}}{\text{tr}\left(\boldsymbol{\Sigma}\right)}\ln{\frac{\boldsymbol{\Sigma}}{\text{tr}\left(\boldsymbol{\Sigma}\right)}}\right)=-\sum_{i=1}^{3}{\frac{s_i^2}{\text{tr}\left(\boldsymbol{\Sigma}\right)}}\ln{\frac{s_i^2}{\text{tr}\left(\boldsymbol{\Sigma}\right)}},
\label{eq:gs_entropy}
\end{gather}where $\text{tr}\left(\boldsymbol{\Sigma}\right)=s_1^2+s_2^2+s_3^2$. These metrics can be used to measure the shape or degree of anisotropy of the Gaussian. In Figure~\ref{fig:spectrum}, we visualize different 3D/2D Gaussians with the same spectral radius, \ie, $\rho\left(\boldsymbol{\Sigma}\right)=16$. 

It is easily proven that when $s_1 = s_2 = s_3$, the condition number is minimized and the spectral entropy is maximized. Furthermore, the eccentricity $e$ of the ellipse described by $\boldsymbol{\Sigma}$ of the 2D Gaussian satisfies $e=\sqrt{1-\frac{1}{\kappa\left(\boldsymbol{\Sigma}\right)}}$, which indicates that a higher condition number corresponds to a sharper shape of the Gaussian as shown in Figure~\ref{fig:spectrum}. And the needle-like artifacts correspond to Gaussians with a low spectral entropy and a high condition number. Please refer to the detailed proofs in the supplementary materials.

\begin{figure*}[t]
\centering
\includegraphics[width=1\textwidth]{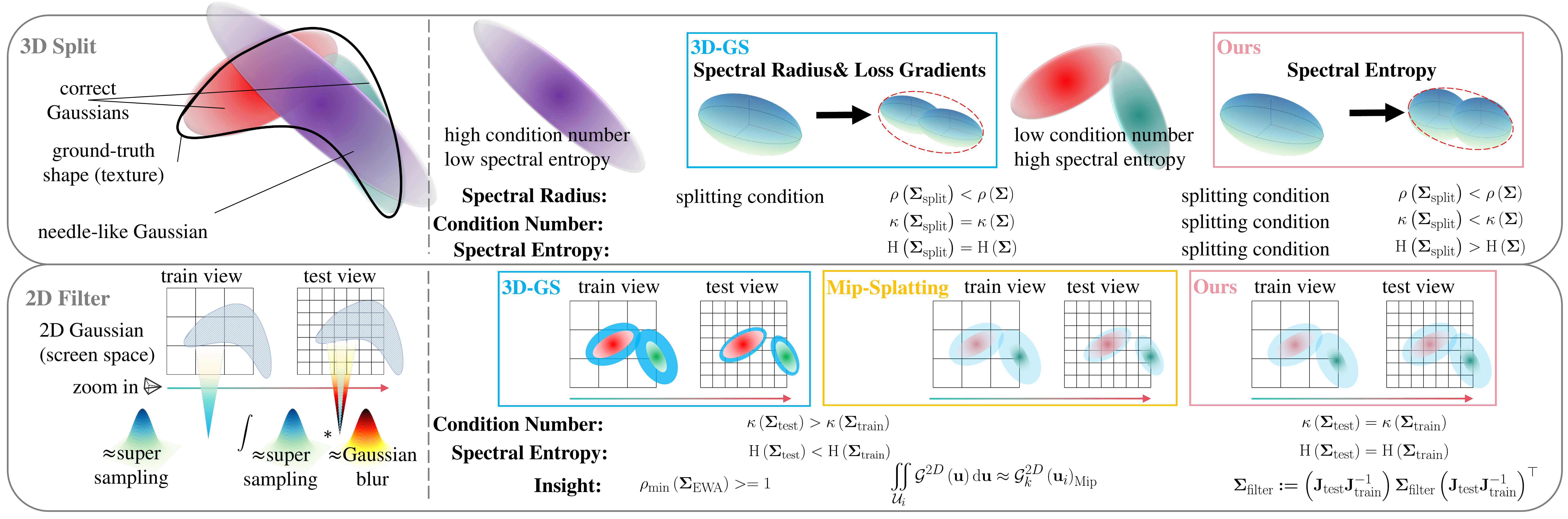}
\caption{Overview of \textcolor{oursred}{\textbf{Spectral-GS}}. \textbf{\textcolor{3dgsblue}{3D Gaussian Splatting (3D-GS)}}~\cite{kerbl20233d}  decides whether to split based on the positional gradients and the spectral radius of the covariance matrix without considering the shape of primitives. We propose the 3D shape-aware splitting strategy based on the spectral analysis \textcolor{gray}{\textbf{(3D Split)}}. In screen space, both the EWA filter~\cite{zwicker2002ewa} of \textbf{\textcolor{3dgsblue}{3D-GS}} which attempts to cover an entire pixel, and the Mip filter of \textbf{\textcolor{mipyellow}{Mip-Splatting}}~\cite{mip_gs} which approximates supersampling, result in a reduction of spectral entropy when zooming in to synthesize novel view. Our view-consistent filter's kernel is not constant to maintain the spectral entropy consistency \textcolor{gray}{\textbf{(2D Filter)}}. } 
\label{fig:overview}
\end{figure*}

\vspace{0.1cm}\noindent{\bf Loss-Sensitivity and Shape-Unawareness in Densification:} In Section~\ref{subsec:3dgs}, we have briefly introduced the optimization and densification in 3D-GS~\cite{kerbl20233d}. Here, we further analyze the spectra of the Gaussians involved in the densification. As shown in Figure~\ref{fig:issue_split}, 3D-GS employs densification to grow the quantity of Gaussians when the loss gradient $\nabla_{\boldsymbol{\mu}_{\text{proj}}}\mathcal{L}$ exceed a certain threshold $\tau_{\text{loss}}$. While this approach effectively addresses issues of “over-reconstruction” and “under-reconstruction”, it is highly sensitive to the design of the loss function $\mathcal{L}$ and the chosen threshold $\tau_{\text{loss}}$. Specifically, when elongated Gaussians with low spectral entropy can fit high-frequency textures or geometry with small loss in the training views, the densification mechanism is not triggered. This can lead to needle-like artifacts or high-frequency over-blurriness. 
Furthermore, even when splitting, the spectral entropy and the condition number of the Gaussians remain consistent with that before the split and does not significantly alleviate needle-like artifacts, \ie, $\rho\left(\boldsymbol{\Sigma}_{\text{split}}\right)=\frac{1}{k^2}\rho\left(\boldsymbol{\Sigma}\right),\kappa\left(\boldsymbol{\Sigma}_{\text{split}}\right)=\kappa\left(\boldsymbol{\Sigma}\right),\text{H}\left(\boldsymbol{\Sigma}_{\text{split}}\right)=\text{H}\left(\boldsymbol{\Sigma}\right)$ (\textcolor{red}{red} text and lines in Figure~\ref{fig:issue_split}).



\vspace{0.1cm}\noindent{\bf View-Inconsistency in Filtering:} In Section~\ref{subsec:3dgs}, we have provided a brief introduction to the EWA filter~\cite{zwicker2002ewa} in 3D-GS and the 2D Mip filter in Mip-Splatting~\cite{mip_gs}. From Equation~\ref{eq:ewa_filter} and~\ref{eq:mip_filter}, 
 we observe that the two filters have identical covariance matrices $\boldsymbol{\Sigma}_{\text{filter}}=\boldsymbol{\Sigma}_{\text{proj}}+s\mathbf{I}$, differing only in the opacity term, \ie, $o\ne o\sqrt{\frac{\left|\boldsymbol{\Sigma}_{\text{proj}}\right|}{\left|\boldsymbol{\Sigma}_{\text{proj}}+s\mathbf{I}\right| }}$. However, when zooming in, the Jacobian matrix $\mathbf{J}$ changes, leading to a change in the projected covariance matrix $\boldsymbol{\Sigma}_{\text{proj}}$ while the filter kernel $s\mathbf{I}$ remains constant. This causes variations in the condition number of 2D Gaussians after optimization:
 \begin{gather}
\kappa\left(\boldsymbol{\Sigma}_{\text{train}}\right)=\frac{\rho\left(\mathbf{J}_{\text{train}}\boldsymbol{\Sigma}^{'}\mathbf{J}_{\text{train}}^{\top}\right)+s}{\rho_{\min}\left(\mathbf{J}_{\text{train}}\boldsymbol{\Sigma}^{'}\mathbf{J}_{\text{train}}^{\top}\right)+s}\ne\kappa\left(\boldsymbol{\Sigma}_{\text{test}}\right)
 \end{gather} 
where $\boldsymbol{\Sigma}^{'}=\mathbf{W}\boldsymbol{\Sigma}\mathbf{W}^{\top}$ is the covariance matrix in the camera space and $\boldsymbol{\Sigma}_{\text{train}}=\mathbf{J}_{\text{train}}\boldsymbol{\Sigma}^{'}\mathbf{J}_{\text{train}}^{\top}+s\mathbf{I}, \boldsymbol{\Sigma}_{\text{test}}=\mathbf{J}_{\text{test}}\boldsymbol{\Sigma}^{'}\mathbf{J}_{\text{test}}^{\top}+s\mathbf{I}$ denote the covariance matrices during training and testing, respectively. We have derived and visualized the variation curve of $\kappa\left(\boldsymbol{\Sigma}_{\text{filter}}\right)$ during the zoom-in process. As shown in the upper part of Figure~\ref{fig:zoomin}, this function increases as the camera zooms in, \ie, as the $\frac{\text{focal length}}{\text{depth}}$ increases. Please refer to the supplementary materials for detailed derivations. 

Due to the view-inconsistency in filtering, needle-like artifacts become more pronounced when zooming in or when the camera moves closer to the object, as illustrated by the rendering results in the lower portion of Figure~\ref{fig:zoomin}. 



\section{Spectral-GS}

Based on the spectral analysis for 3D-GS~\cite{kerbl20233d} and Mip-Splatting~\cite{mip_gs} in Section~\ref{sec:spectral_analysis_3dgs}, we propose the 3D shape-aware splitting in Section~\ref{subsec:shape_splitting} and the 2D view-consistent filtering in Section~\ref{subsec:view_filtering}, respectively, to address loss sensitivity and shape unawareness in densification, as well as view inconsistency in filtering.  The
overview of our method Spectral-GS is illustrated in Figure~\ref{fig:overview}.

\subsection{3D Shape-Aware Splitting}
\label{subsec:shape_splitting}

We propose the 3D shape-aware splitting to introduce shape-awareness into the optimization process. As shown in \textcolor{oursred}{\textbf{Ours}} of \textcolor{gray}{\textbf{3D Split}} in Figure~\ref{fig:overview}, the splitting condition of our strategy is based on the spectral entropy of 3D Gaussians $\text{H}\left(\boldsymbol{\Sigma}\right)$. When the spectral entropy exceeds a certain threshold $\tau_{\text{spectral}}$ and may exhibit visually needle-like artifacts, $K$ points are sampled based on the probability density function (PDF) of the old Gaussian distribution $\mathcal{G}^{3D}\left(\mathbf{x};\boldsymbol{\mu},\boldsymbol{\Sigma}\right)$. The new Gaussian mixture distributions $\sum\limits_{K}\mathcal{G}^{3D}\left(\mathbf{x};\boldsymbol{\mu}_{\text{split}},\boldsymbol{\Sigma}_{\text{split}}\right)$ aim to fit the old Gaussian distribution $\mathcal{G}^{3D}\left(\mathbf{x};\boldsymbol{\mu},\boldsymbol{\Sigma}\right)$ as closely as possible, preserving high-frequency while increasing spectral entropy. Specifically, the reduction factor of the covariance matrix is not isotropic but anisotropic. The greedy algorithm reduces the spectral radius by the maximum extent:
\begin{gather}
\boldsymbol{\Sigma}_{\text{split}}=\mathbf{R}\text{diag}\left(\frac{1}{k_1^2} s_1^2,
\frac{1}{k_2^2} s_2^2,
\frac{1}{k_3^2} s_3^2\right)\mathbf{R}^{\top},\\
k_i= k\cdot\mathbbm{1}\left\{s_i^2=\rho\left(\boldsymbol{\Sigma}\right)\right\}+k_0,
\end{gather} where $\mathbbm{1}\left\{\cdot\right\}$ is an indicator function and $k>0,k_0\ge1$. Additionally, to ensure that the condition number after splitting does not exceed that before splitting, the following condition must be satisfied:
\begin{equation}
\label{eq:k_condition}
    k < - k_{0} + \frac{k_0\rho^{\frac{3}{2}}\left(\boldsymbol{\Sigma}\right)}{\sqrt{\left|\boldsymbol{\Sigma}\right|}}\text{.}
\end{equation} Please refer to the supplementary materials for more details.

\subsection{2D View-Consistent Filtering}
\label{subsec:view_filtering}

For novel view synthesis, the level of scene detail is determined by the resolution of the training images. The finer details revealed by zooming in actually correspond to views that are not seen in the training set. Due to view-inconsistent filtering, 3D-GS~\cite{kerbl20233d} and Mip-Splatting~\cite{mip_gs} produce pronounced artifacts when zooming in. We propose the 2D view-consistent filtering, which combining a convolution that approximates supersampling with a Gaussian blur that approximates interpolation, as shown in \textcolor{oursred}{\textbf{Ours}} of \textcolor{gray}{\textbf{2D Filter}} in Figure~\ref{fig:overview}. We first tend to approximate the integral of projected 2D Gaussians within each pixel window area with the 2D Gaussian filter:
\begin{gather}
\mathcal{G}^{2D}_{k}\left(\mathbf{u}\right)_{\text{Box}}=\left(\mathcal{G}^{2D}\otimes\mathcal{B}\right)\left(\mathbf{u}\right)=\iint\limits_{\mathcal{U}}{\mathcal{G}^{2D}}\left(\mathbf{u}\right)\mathrm{d}\mathbf{u},\\
\mathcal{G}^{2D}_{k}\left(\mathbf{u}\right)_{\text{Box}}\approx o\sqrt{\frac{\left|\boldsymbol{\Sigma}_{\text{proj}}\right|}{\left|\boldsymbol{\Sigma}_{\text{proj}}+s\mathbf{I}\right| }}e^{-\frac{1}{2}\left(\mathbf{u}-\boldsymbol{\mu}_{\text{proj}}\right)^{\top}
{\left(\boldsymbol{\Sigma}_{\text{proj}}+s\mathbf{I}\right)}^{-1}
\left(\mathbf{u}-\boldsymbol{\mu}_{\text{proj}}\right)},
\end{gather} where $\mathcal{B}$ denotes the 2D box filter. Subsequently, we introduce the Gaussian blur to achieve view-consistency of the condition number, with the size of the convolution kernel given by:
\begin{equation}
\boldsymbol{\Sigma}_{\text{blur}}=\left(\mathbf{J}_{\text{test}}\mathbf{J}_{\text{train}}^{-1}\right)s\mathbf{I}\left(\mathbf{J}_{\text{test}}\mathbf{J}_{\text{train}}^{-1}\right)^{\top}-s\mathbf{I}.
\end{equation} Note that the matrix $\mathbf{J}_{\text{train}} \in \mathbb{R}^{2 \times 3}$ is not a full-rank square matrix; $ \mathbf{J}_{\text{train}}^{-1} \in \mathbb{R}^{3 \times 2} $ represents a left/right inverse inverse of matrix $\mathbf{J}_{\text{train}}$. Since convolving two Gaussians with
covariance matrices $\boldsymbol{\Sigma}_1, \boldsymbol{\Sigma}_2$ results in another Gaussian
with variance $\boldsymbol{\Sigma}_1 + \boldsymbol{\Sigma}_2$, we obtain the result of combining the 2D box filter with Gaussian blur $\mathcal{G}^{blur}$ as follows:
\begin{gather}
\begin{split}
\mathcal{G}_k^{2D}\left(\mathbf{u}\right)_{\text{filter}}&\approx\left(\mathcal{G}^{2D}\otimes\mathcal{B}\otimes\mathcal{G}^{blur}\right)\left(\mathbf{u}\right)\\&=o_{\text{filter}}e^{-\frac{1}{2}\left(\mathbf{u}-\boldsymbol{\mu}_{\text{proj}}\right)^{\top}
{\boldsymbol{\Sigma}_{\text{filter}}}^{-1}
\left(\mathbf{u}-\boldsymbol{\mu}_{\text{proj}}\right)}
\end{split},\\
o_{\text{filter}}=o\sqrt{\frac{\left|\boldsymbol{\Sigma}_{\text{proj}}\right|}{\left|\boldsymbol{\Sigma}_{\text{proj}}+\left(\mathbf{J}_{\text{test}}\mathbf{J}_{\text{train}}^{-1}\right)s\mathbf{I}\left(\mathbf{J}_{\text{test}}\mathbf{J}_{\text{train}}^{-1}\right)^{\top}\right| }},\\
\label{eq:spectral_filter}
\boldsymbol{\Sigma}_{\text{filter}}=\boldsymbol{\Sigma}_{\text{proj}}+\left(\mathbf{J}_{\text{test}}\mathbf{J}_{\text{train}}^{-1}\right)s\mathbf{I}\left(\mathbf{J}_{\text{test}}\mathbf{J}_{\text{train}}^{-1}\right)^{\top}.
\end{gather} Compared to the 2D Mip filter, our 2D view-consistent filter does not use a constant kernel but instead employs a view-adaptive kernel. We can also approximate Equation~\ref{eq:spectral_filter} using the filter kernel function $s\left(\text{focal length}, \text{depth}\right)=s_0\frac{\text{focal length}^2}{\text{depth}^2}$, where $s_0$ is a constant. Please
refer to the detailed proofs in the supplementary materials.



\begin{figure}[!t]
\centering
\includegraphics[width=1\linewidth]{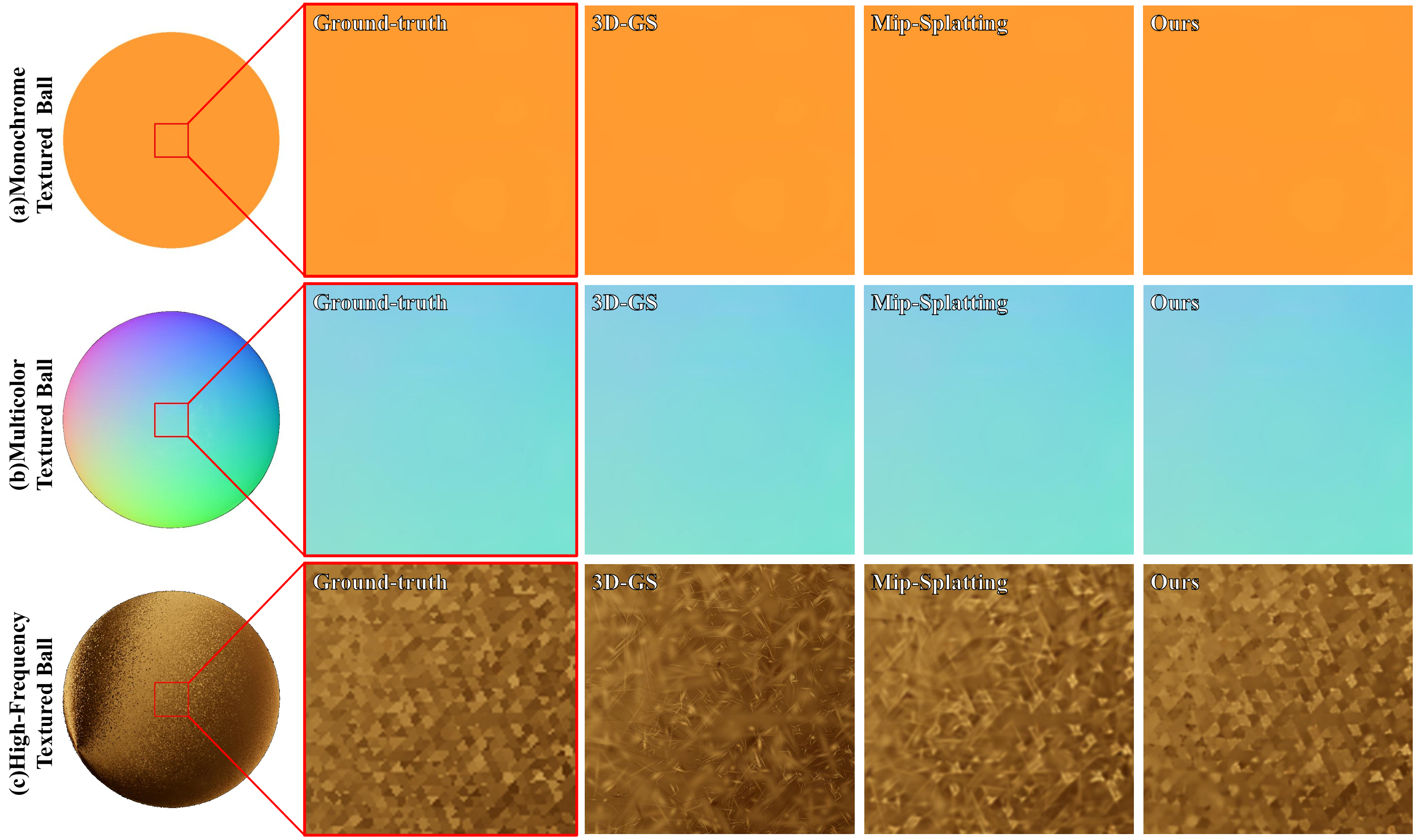}
\caption{We conduct experiments on \textsc{Ball} with identical geometry but different textures. From top to bottom, they are the monochrome textured \textbf{(a)}, multicolor textured \textbf{(b)}, and high-frequency textured \textbf{(c)}.} 
\label{fig:impacts_of_freq}
\end{figure}

\section{Experiments}

To validate the effectiveness of Spectral-GS derived from spectral analysis, we conduct a series of experiments, comparing it with the original 3D-GS~\cite{kerbl20233d} and some current state-of-the-art methods. In these experiments, Analytic-Splatting+3D Filter refers to Analytic-Splatting~\cite{liang2024analytic} with the 3D smoothing filter from Mip-Splatting~\cite{mip_gs}.

\subsection{Implementation}

We implement Spectral-GS based on the PyTorch framework in 3D-GS~\cite{kerbl20233d}. We use the default parameters of 3D-GS to maintain consistency with the original 3D-GS. For our approach, we empirically set the threshold $\tau_{\text{spectral}}$ in 3D shape-aware splitting to $0.5$, with $k = 0.6$, $k_0 = 1$ and $K=2$, and $s_0$ in 2D view-consistent filtering to $0.1$.

\subsection{Datasets}

We test our algorithm on a total of 12 scenes which are commonly used in NeRF~\cite{mildenhall2020nerf} and 3D-GS~\cite{kerbl20233d}.

\vspace{0.1cm}\noindent{\bf Synthetic Scenes~\cite{mildenhall2020nerf, verbin2022ref}:} In particular, we evaluate our approach on a total of six synthetic scenes. \textsc{Hotdog}, \textsc{Chair}, \textsc{Ship}, \textsc{Lego}, and \textsc{Materials} are from the \textit{Blender dataset}~\cite{mildenhall2020nerf}, while \textsc{Ball} is from the \textit{Shiny Blender dataset}~\cite{verbin2022ref} but  with modified textures.

\vspace{0.1cm}\noindent{\bf Real Scenes~\cite{barron2022mip,knapitsch2017tanks, hedman2018deep, jensen2014large}:} Additionally, we test our method on a total of six real scenes. \textsc{Truck}, \textsc{Playroom} and \textsc{Flowers} are from the \textit{Tanks Templates}~\cite{knapitsch2017tanks}, \textit{Deep Blending}~\cite{hedman2018deep} and the \textit{Mip-NeRF360}~\cite{barron2022mip} datasets, respectively. \textsc{Tripod}, \textsc{Stone} and \textsc{Pillow} are \textit{High-frequency Spectrum} dataset captured by us. 

\subsection{Results}

\vspace{0.1cm}\noindent{\bf Quantitative comparisons:} We adopt a train/test split for real datasets following the methodology proposed by 3D-GS~\cite{kerbl20233d}. Standard metrics such as PSNR, LPIPS~\cite{lpips}, and SSIM are employed for evaluation. We report quantiative results in Table~\ref{tab:comparison}. Additionally, to verify the correlation between the spectral entropy of 3D Gaussians after optimization $\text{H}\left(\boldsymbol{\Sigma}\right)$ and the quality of novel view synthesis, we also provide the spectral entropy metric for each scene in Table~\ref{tab:comparison}. We employ the setting involving renderings at various focal lengths (\ie $1\times$, $2\times$, $4\times$, $8\times$) to mimic zoom-in effects. Table~\ref{tab:comparison} demonstrates that our spectral analysis-based method effectively increases the spectral entropy of scenes, thereby enhancing images' quality.

\begin{figure*}[t]
\centering
\includegraphics[width=\textwidth]{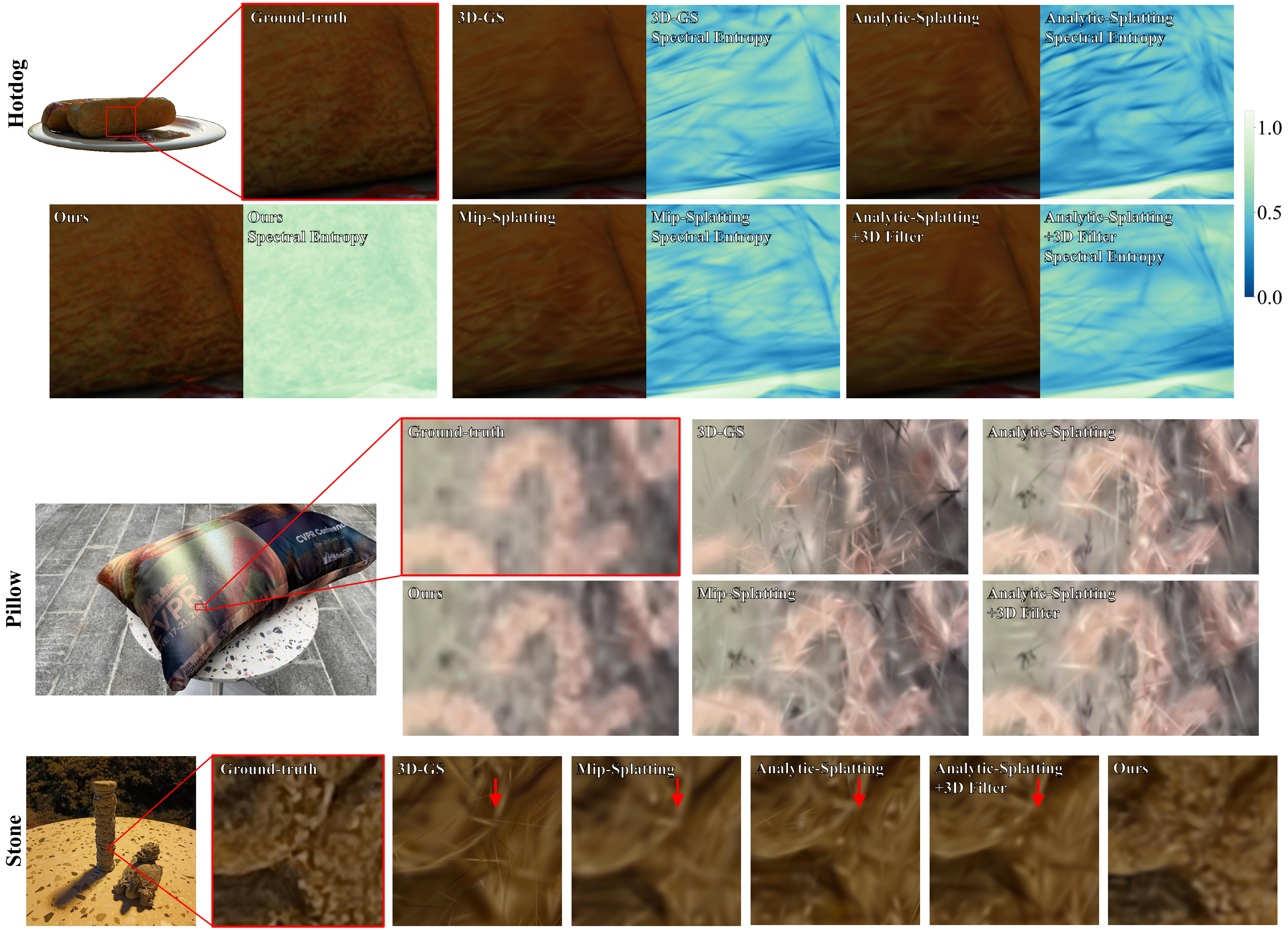}
\caption{We show comparisons of our method to previous methods and the corresponding ground truth images from held-out test views. Additionally, We visualize the spectral entropy maps of 3D Gaussians after optimization. Bluer regions indicate lower spectral entropy, with more needle-like degraded Gaussians, while greener regions represent higher spectral entropy, without noticeable needle-like artifacts.} 
\label{fig:cmp}

\end{figure*}

\vspace{0.1cm}\noindent{\bf Qualitative comparisons:} As illustrated in Figure~\ref{fig:cmp}, it can be observed that our method is capable of generating more realistic details, with fewer needle-like artifacts compared to 3D-GS~\cite{kerbl20233d}, Mip-Splatting~\cite{mip_gs} and Analytic-Splatting~\cite{liang2024analytic}. 
And this is precisely the superiority brought about by our spectral analysis-based method, which results in higher spectral entropy.

\subsection{Discussions}
\label{subsec:discuss}

\begin{figure}[t]
\centering
\includegraphics[width=1\linewidth]{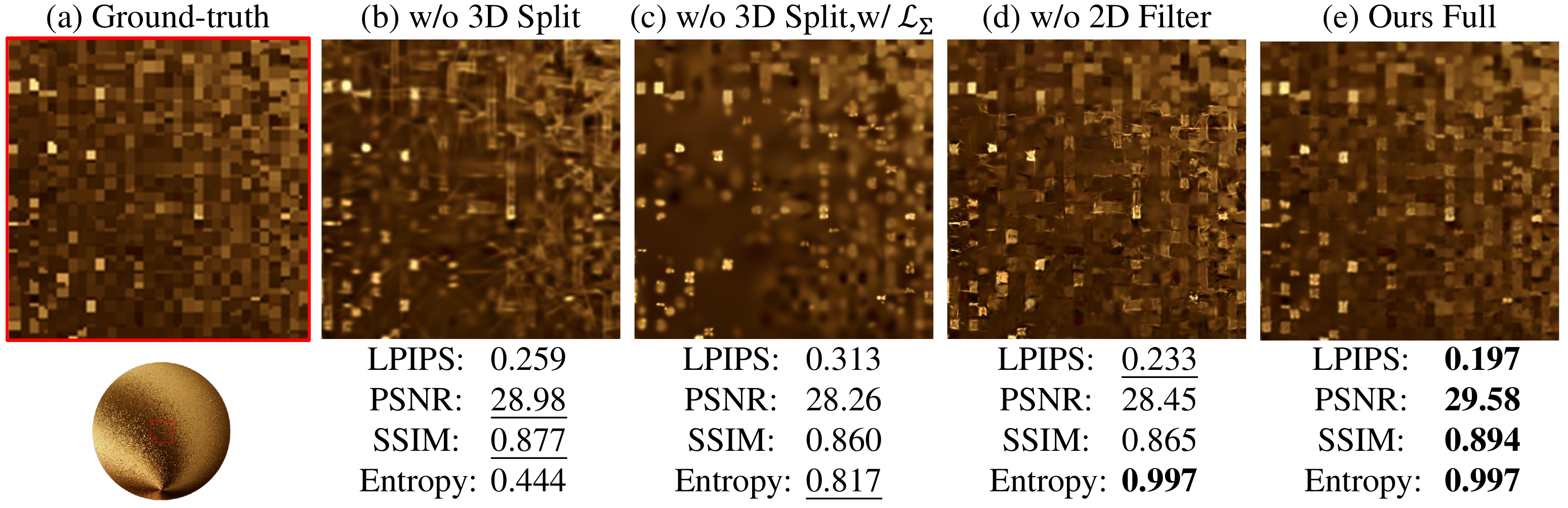}
\caption{Illustrations of the ablation study on \textsc{Ball}. From left to right: ground truth \textbf{(a)}, our method without 3D split \textbf{(b)}, our method without 3D split but with  $\mathcal{L}_{\boldsymbol{\Sigma}}$ \textbf{(c)}, our method without 2D filter \textbf{(d)}, and the full version of our method \textbf{(e)}. The corresponding metrics are provided below the image.} 
\label{fig:ablation}
\end{figure}

\begin{figure}[t]
\centering
\includegraphics[width=1\linewidth]{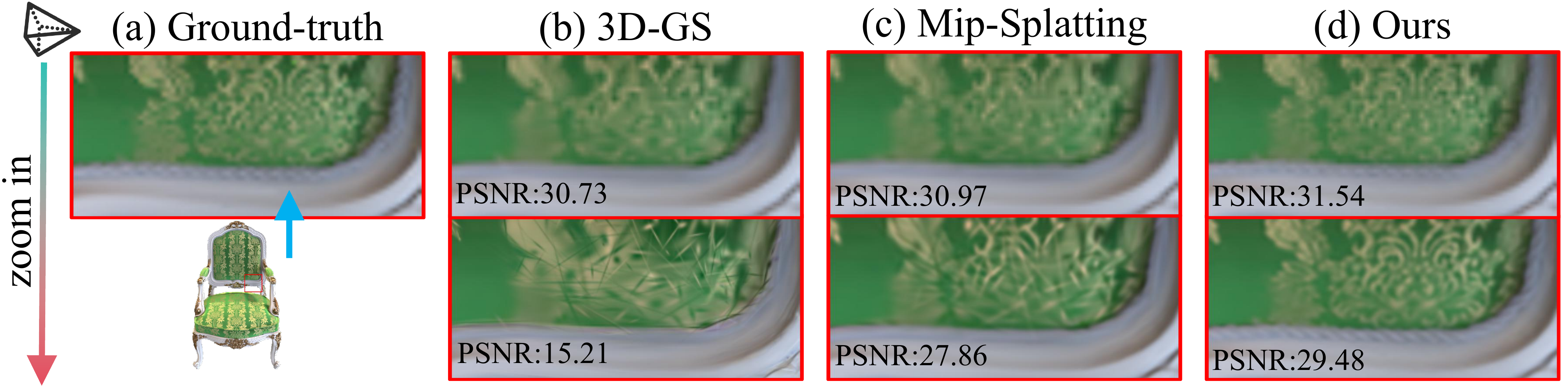}
\caption{We present results from different methods at various sampling rates (focal lengths) on \textsc{Chair}. The images are ordered from top to bottom, corresponding to the transition from the training focal length to larger focal lengths. Each method includes a PSNR metric in the lower-left corner.} 
\label{fig:impacts_of_focal}
\end{figure}

\begin{table*}[!t]
\centering
\caption{
Quantitative evaluation of our method compared to previous work across 12 scenes.}
\label{tab:comparison}
  	\small
   	\scalebox{0.80}{
\begin{tabular}{c|c|cccccc|cccccc}
                                   &                              & \multicolumn{6}{c}{Synthetic Dataset~\cite{mildenhall2020nerf,verbin2022ref}}                                                                                                                                                         & \multicolumn{6}{c}{Real Dataset~\cite{knapitsch2017tanks, hedman2018deep, barron2022mip}}                                                                                                                                                              \\
\multirow{-2}{*}{Metric}           & \multirow{-2}{*}{Method}     & Hotdog                        & Chair                         & Ship                          & Lego                          & Materials                     & Ball                          & Truck                         & Playroom                      & Flowers                       & Tripod                        & Stone                           & Pillow                       \\\hline
                                   & 3D-GS~\cite{kerbl20233d}                        & \cellcolor[HTML]{FFFFFF}0.204 & \cellcolor[HTML]{FFFFFF}0.213 & \cellcolor[HTML]{FFFFFF}0.310 & \cellcolor[HTML]{FFFFFF}0.189 & \cellcolor[HTML]{FFFFFF}0.141 & \cellcolor[HTML]{FFFFFF}0.377 & \cellcolor[HTML]{FFFFFF}0.221 & \cellcolor[HTML]{FFFFFF}0.215 & \cellcolor[HTML]{FFFFFF}0.366 & \cellcolor[HTML]{FFFFFF}0.241 & 0.280 & \cellcolor[HTML]{FFFFFF}0.292 \\
                                   & Mip-Splatting~\cite{mip_gs}                & \cellcolor[HTML]{FFD9B2}0.126 & \cellcolor[HTML]{FFFFB2}0.114 & \cellcolor[HTML]{FFD9B2}0.216 & \cellcolor[HTML]{FFD9B2}0.111 & \cellcolor[HTML]{FFD9B2}0.113 & \cellcolor[HTML]{FFD9B2}0.285 & \cellcolor[HTML]{FFD9B2}0.166 & \cellcolor[HTML]{FFD9B2}0.201 & \cellcolor[HTML]{FFD9B2}0.315 & \cellcolor[HTML]{FFD9B2}0.222 & \cellcolor[HTML]{FFD9B2}0.230 & \cellcolor[HTML]{FFD9B2}0.275 \\
                                   & Analytic-Splatting~\cite{liang2024analytic}           & 0.165                         & 0.173                         & 0.280                         & 0.150                         & 0.141                         & \cellcolor[HTML]{FFFFB2}0.293 & \cellcolor[HTML]{FFFFB2}0.176 & \cellcolor[HTML]{FFFFB2}0.205 & \cellcolor[HTML]{FFFFB2}0.342 & 0.228                         & 0.281                         & 0.289 \\
                                   & Analytic-Splatting+3D Filter~\cite{liang2024analytic} & \cellcolor[HTML]{FFFFB2}0.133 & \cellcolor[HTML]{FFD9B2}0.112 & \cellcolor[HTML]{FFFFB2}0.219 & \cellcolor[HTML]{FFFFB2}0.119 & \cellcolor[HTML]{FFFFB2}0.115 & 0.294                         & \cellcolor[HTML]{FFFFB2}0.176 & \cellcolor[HTML]{FFFFB2}0.205 & \cellcolor[HTML]{FFFFB2}0.342 & \cellcolor[HTML]{FFFFB2}0.226 & \cellcolor[HTML]{FFFFB2}0.274                        & \cellcolor[HTML]{FFFFB2}0.287                         \\
\multirow{-5}{*}{LPIPS$\downarrow$}            & Ours                         & \cellcolor[HTML]{FFB2B2}0.099 & \cellcolor[HTML]{FFB2B2}0.098 & \cellcolor[HTML]{FFB2B2}0.196 & \cellcolor[HTML]{FFB2B2}0.098 & \cellcolor[HTML]{FFB2B2}0.100 & \cellcolor[HTML]{FFB2B2}0.197 & \cellcolor[HTML]{FFB2B2}0.129 & \cellcolor[HTML]{FFB2B2}0.186 & \cellcolor[HTML]{FFB2B2}0.307 & \cellcolor[HTML]{FFB2B2}0.214 & \cellcolor[HTML]{FFB2B2}0.217 & \cellcolor[HTML]{FFB2B2}0.266 \\ \hline
                                   & 3D-GS~\cite{kerbl20233d}                        & 29.17                         & 23.10                         & 23.41                         & 28.23                         & 27.25                         & 23.85                         & 25.16                         & 31.12                         & 18.97                         & 21.92                         & 27.13 & 25.28                         \\
                                   & Mip-Splatting~\cite{mip_gs}                & \cellcolor[HTML]{FFD9B2}33.85 & \cellcolor[HTML]{FFFFB2}30.75 & \cellcolor[HTML]{FFFFB2}28.44 & \cellcolor[HTML]{FFD9B2}31.86 & \cellcolor[HTML]{FFD9B2}29.18 & 29.03                         & \cellcolor[HTML]{FFD9B2}26.59 & \cellcolor[HTML]{FFFFB2}31.85 & \cellcolor[HTML]{FFD9B2}21.90 & \cellcolor[HTML]{FFD9B2}23.49 & 29.48 & 
                                   \cellcolor[HTML]{FFFFB2} 25.94 \\
                                   & Analytic-Splatting~\cite{liang2024analytic}           & 32.99                         & 29.27                         & 26.93                         & 31.44                         & 28.75                         & \cellcolor[HTML]{FFFFB2}29.08 & \cellcolor[HTML]{FFFFB2}26.57 & \cellcolor[HTML]{FFB2B2}31.92 & \cellcolor[HTML]{FFFFB2}21.64 & \cellcolor[HTML]{FFFFB2}23.15 & \cellcolor[HTML]{FFD9B2}29.65                         & 25.90 \\
                                   & Analytic-Splatting+3D Filter~\cite{liang2024analytic} & \cellcolor[HTML]{FFFFB2}33.73 & \cellcolor[HTML]{FFD9B2}30.92 & \cellcolor[HTML]{FFD9B2}28.46 & \cellcolor[HTML]{FFFFB2}31.47 & \cellcolor[HTML]{FFFFB2}29.09 & \cellcolor[HTML]{FFD9B2}29.10 & 26.53                         & 31.77                         & \cellcolor[HTML]{FFFFB2}21.64 & 23.14                         & \cellcolor[HTML]{FFD9B2}29.65                         & \cellcolor[HTML]{FFD9B2}25.96                         \\
\multirow{-5}{*}{PSNR$\uparrow$}             & Ours                         & \cellcolor[HTML]{FFB2B2}34.16 & \cellcolor[HTML]{FFB2B2}31.51 & \cellcolor[HTML]{FFB2B2}28.79 & \cellcolor[HTML]{FFB2B2}32.29 & \cellcolor[HTML]{FFB2B2}29.43 & \cellcolor[HTML]{FFB2B2}29.58 & \cellcolor[HTML]{FFB2B2}26.86 & \cellcolor[HTML]{FFB2B2}31.92 & \cellcolor[HTML]{FFB2B2}22.11 & \cellcolor[HTML]{FFB2B2}24.37 & \cellcolor[HTML]{FFB2B2}30.13 & \cellcolor[HTML]{FFB2B2}26.34 \\ \hline
                                   & 3D-GS~\cite{kerbl20233d}                        & 0.889                         & 0.795                         & 0.718                         & 0.887                         & 0.914                         & 0.750                         & 0.866                         & 0.923                         & 0.542                         & 0.725                         & 0.854 & 0.709                         \\
                                   & Mip-Splatting~\cite{mip_gs}                & \cellcolor[HTML]{FFD9B2}0.952 & \cellcolor[HTML]{FFD9B2}0.943 & \cellcolor[HTML]{FFFFB2}0.864 & \cellcolor[HTML]{FFD9B2}0.944 & \cellcolor[HTML]{FFD9B2}0.936 & 0.873                         & \cellcolor[HTML]{FFD9B2}0.893 & \cellcolor[HTML]{FFD9B2}0.928 & \cellcolor[HTML]{FFD9B2}0.637 & \cellcolor[HTML]{FFD9B2}0.767 & \cellcolor[HTML]{FFFFB2}0.892                         & 
                                   \cellcolor[HTML]{FFD9B2}0.720 \\
                                   & Analytic-Splatting~\cite{liang2024analytic}           & 0.931                         & 0.864                         & 0.796                         & 0.929                         & 0.925                         & \cellcolor[HTML]{FFD9B2}0.875 & \cellcolor[HTML]{FFFFB2}0.891 & 0.927                         & \cellcolor[HTML]{FFFFB2}0.620 & 0.754                         & 0.784 & 0.718 \\
                                   & Analytic-Splatting+3D Filter~\cite{liang2024analytic} & \cellcolor[HTML]{FFD9B2}0.952 & \cellcolor[HTML]{FFFFB2}0.934 & \cellcolor[HTML]{FFD9B2}0.865 & \cellcolor[HTML]{FFFFB2}0.940 & \cellcolor[HTML]{FFD9B2}0.936 & \cellcolor[HTML]{FFD9B2}0.875 & \cellcolor[HTML]{FFFFB2}0.891 & \cellcolor[HTML]{FFD9B2}0.928 & \cellcolor[HTML]{FFFFB2}0.620 & \cellcolor[HTML]{FFFFB2}0.756 & \cellcolor[HTML]{FFB2B2}0.919                         & \cellcolor[HTML]{FFFFB2}0.719                         \\
\multirow{-5}{*}{SSIM$\uparrow$}             & Ours                         & \cellcolor[HTML]{FFB2B2}0.956 & \cellcolor[HTML]{FFB2B2}0.945 & \cellcolor[HTML]{FFB2B2}0.866 & \cellcolor[HTML]{FFB2B2}0.947 & \cellcolor[HTML]{FFB2B2}0.937 & \cellcolor[HTML]{FFB2B2}0.894 & \cellcolor[HTML]{FFB2B2}0.907 & \cellcolor[HTML]{FFB2B2}0.930 & \cellcolor[HTML]{FFB2B2}0.644 & \cellcolor[HTML]{FFB2B2}0.773 & \cellcolor[HTML]{FFD9B2}0.914 & \cellcolor[HTML]{FFB2B2}0.735 \\ \hline
                                   & 3D-GS~\cite{kerbl20233d}                        & 0.152                         & 0.083                         & 0.141                         & 0.189                         & 0.242                         & 0.140                         & 0.254                         & 0.276                         & 0.239                         & 0.357                         & 0.351                         & 0.398                         \\
                                   & Mip-Splatting~\cite{mip_gs}                & \cellcolor[HTML]{FFD9B2}0.339 & \cellcolor[HTML]{FFD9B2}0.397 & \cellcolor[HTML]{FFD9B2}0.338 & \cellcolor[HTML]{FFD9B2}0.361 & \cellcolor[HTML]{FFD9B2}0.397 & \cellcolor[HTML]{FFFFB2}0.444 & \cellcolor[HTML]{FFD9B2}0.544 & \cellcolor[HTML]{FFD9B2}0.349 & \cellcolor[HTML]{FFFFB2}0.664 & \cellcolor[HTML]{FFD9B2}0.442 & \cellcolor[HTML]{FFD9B2}0.532                         & \cellcolor[HTML]{FFD9B2}0.489 \\
                                   & Analytic-Splatting~\cite{liang2024analytic}           & 0.234                         & 0.222                         & 0.210                         & 0.241                         & 0.285                         & 0.362                         & 0.377                         & 0.292                         & 0.492                         & 0.416                         & 0.430 & 0.467                         \\
                                   & Analytic-Splatting+3D Filter~\cite{liang2024analytic} & \cellcolor[HTML]{FFFFB2}0.332 & \cellcolor[HTML]{FFFFB2}0.390 & \cellcolor[HTML]{FFFFB2}0.334 & \cellcolor[HTML]{FFFFB2}0.354 & \cellcolor[HTML]{FFD9B2}0.397 & \cellcolor[HTML]{FFD9B2}0.496 & \cellcolor[HTML]{FFFFB2}0.518 & \cellcolor[HTML]{FFFFB2}0.321 & \cellcolor[HTML]{FFD9B2}0.682 & \cellcolor[HTML]{FFFFB2}0.425 & \cellcolor[HTML]{FFFFB2}0.478 & \cellcolor[HTML]{FFFFB2}0.479 \\
\multirow{-5}{*}{Entropy$\uparrow$} & Ours                         & \cellcolor[HTML]{FFB2B2}0.697 & \cellcolor[HTML]{FFB2B2}0.992 & \cellcolor[HTML]{FFB2B2}0.996 & \cellcolor[HTML]{FFB2B2}0.970 & \cellcolor[HTML]{FFB2B2}0.967 & \cellcolor[HTML]{FFB2B2}0.997 & \cellcolor[HTML]{FFB2B2}0.881 & \cellcolor[HTML]{FFB2B2}0.794 & \cellcolor[HTML]{FFB2B2}0.958 & \cellcolor[HTML]{FFB2B2}0.836 & \cellcolor[HTML]{FFB2B2}0.908 & \cellcolor[HTML]{FFB2B2}0.615

\end{tabular}
}
\end{table*}

\vspace{0.1cm}\noindent{\bf Ablation Study:}  We conduct ablation studies in the \textsc{Ball} scene, as shown in Figure~\ref{fig:ablation}. The observed needle-like artifacts are composed of two parts: those inherent to the 3D scene and those generated by the rendering algorithm. The artifacts inherent to the scene are addressed by regulating the 3D Gaussians' spectral entropy through our 3D splitting \textbf{(b)(e)}, while the artifacts caused by rendering are resolved by maintaining condition number consistency via our 2D filtering \textbf{(d)(e)}. Moreover, 3D splitting, which does not rely on loss gradients, enhances the representation of high-frequency details in the scene. Recall that 3D-GS~\cite{kerbl20233d} is sensitive to loss and lacks shape-awareness. Another idea is to introduce a regularization term $\mathcal{L}_{\boldsymbol{\Sigma}}$ into the loss function that accounts for 3D Gaussians' shape:
\begin{equation}
    \mathcal{L}_{\text{naive}} = (1-\lambda_1)\mathcal{L}_1 + \lambda_1 \mathcal{L}_{\text{D-SSIM}} + \lambda_2 \mathcal{L}_{\boldsymbol{\Sigma}}.  
\end{equation} However, although this regularization term is able to effectively constrain and optimize the shape of Gaussians ($\text{H}\left(\boldsymbol{\Sigma}\right)=0.817>0.444$), it does not affect densification since the gradient with respect to $\boldsymbol{\mu}_{\text{proj}}$ is zero:
\begin{gather}
\Vert\nabla_{\boldsymbol{\mu}_{\text{proj}}}\mathcal{L}_{\boldsymbol{\Sigma}}\Vert_F=0,\quad\Vert\nabla_{\boldsymbol{\Sigma}}\mathcal{L}_{\boldsymbol{\Sigma}}\Vert_F\ne 0,\\
\Vert\nabla_{\boldsymbol{\mu}_{\text{proj}}}\mathcal{L}_{\text{naive}}\Vert_F=\Vert\nabla_{\boldsymbol{\mu}_{\text{proj}}}\mathcal{L}\Vert_F,
\end{gather}
which can lead to loss of high-frequency details \textbf{(c)(e)}. 


\vspace{0.1cm}\noindent{\bf Relationship with Frequency:} It is worth noting that Mip-Splatting~\cite{mip_gs} outperforms 3D-GS~\cite{kerbl20233d} in both spectral entropy and image quality metrics across all scenes, suggesting a relationship between the 3D smoothing filter based on frequency analysis and ours spectral analysis. Firstly, it is straightforward to prove that the 3D smoothing filter in Mip-Splatting increases the spectral entropy of the original Gaussian. Secondly, we know that Gaussian functions are closed under the Fourier transform (FT) ~\cite{nussbaumer1982fast} while the covariance in the frequency domain is the inverse in the spatial domain (scaled by a coefficient):
\begin{equation}
    \mathcal{F}\left\{\mathcal{G}\left(\mathbf{x}\right)\right\}=e^{-2\pi^2\boldsymbol{\omega}^{\top}\boldsymbol{\Sigma}\boldsymbol{\omega}}
\end{equation} where $\mathcal{F}$ denotes the Fourier transform. The eigenvalues of the covariance matrix determine the bandwidth of the frequency spectrum. Larger eigenvalues result in a narrower spectrum in the corresponding direction, while smaller eigenvalues lead to a wider spectrum in that direction. Therefore, the spectral analysis of our covariance matrix is equivalent and unified with the frequency spectrum analysis of the Gaussian. We conduct experiments with fixed geometry on \textsc{Ball} with different textures to demonstrate that needle-like artifacts predominantly occur in high-frequency scenes, as shown in Figure~\ref{fig:impacts_of_freq}. And our method effectively addresses the challenge of representing high-frequency details with Gaussians without artifacts.

\vspace{0.1cm}\noindent{\bf Impacts of Zooming in:} To further validate the robustness of our method during the zoom-in process, we conduct a series of experiments. As shown in Figure~\ref{fig:impacts_of_focal}, these methods synthesize novel views nearly identical to the ground-truth at the training view resolution, although textures from 3D-GS and Mip-Splatting are blurred at the blue arrow location due to their naive splitting algorithm. However, when increasing the focal length and sampling rate, 3D-GS exhibits pronounced needle-like artifacts that significantly degrade rendering quality and PSNR. While Mip-Splatting shows less degradation due to its use of a series of filters, Gaussian shapes still become sharper. In contrast, we maintain high quality similar to lower sampling rates, without needle-like Gaussians or loss of high-frequency details, through our 2D view-consistent filtering.

\vspace{0.1cm}\noindent{\bf Limitation:} Since our method does not introduce additional priors, such as image super-resolution networks~\cite{dong2015image}, the resolution of novel views depends on the input images. Additionally, our method includes hyperparameters similar to 3D-GS~\cite{kerbl20233d} and Mip-Splatting~\cite{mip_gs}, so the results are influenced by these hyperparameters.
\section{Conclusion}

We propose Spectral-GS, a modification to 3D-GS, which introduces \textit{3D scale-aware splitting} and \textit{2D view-consistent filtering} strategies, based on our spectral analysis, to achieve needle-like-alias-free rendering at arbitrary close-up or zoomed-in view.  Our splitting strategy effectively regularizes
needle-like Gaussians and increase the spectral entropy, enhancing the high-frequency de-
tails representation for 3D-GS and mitigating needle-like
artifacts. And the 2D view-consistent filter combines a convolution that approximates supersampling with a Gaussian blur that approximates interpolation to resolve needle-like artifacts caused by view-inconsistency. Our experimental results validate accuracy of our analysis and effectiveness of our method.

{
    \small
    \bibliographystyle{ieeenat_fullname}
    \bibliography{main}
}


 \clearpage
\setcounter{page}{1}
\maketitlesupplementary

\appendix

\section{Proofs in Spectral Analysis of Gaussians}

\subsection{Spectral Analysis of Matrices}
\label{subsec:spectral_analysis}

Mathematically, the spectrum of a matrix refers to the set of its eigenvalues~\cite{golub2013matrix, zill2020advanced, eisenbud2013commutative}.

\vspace{0.1cm}\noindent{\bf Eigenvalue (Spectrum):} A matrix $\mathbf{A}\in \mathbb{R}^{N\times N}$ can be eigendecomposed as follows:
\begin{equation}
\mathbf{A}=\mathbf{Q}\boldsymbol{\Lambda}\mathbf{Q}^{-1}
\end{equation} where $\boldsymbol{\Lambda}=\text{diag}\left(\begin{matrix}\lambda_1,\lambda_2,\ldots,\lambda_N\end{matrix}\right)$ is the diagonal matrix whose diagonal elements are the corresponding eigenvalues, $\boldsymbol{\Lambda}_{ii} = \lambda_i$. The trace of $\mathbf{A}$, denoted $\text{tr}\left(\mathbf{A}\right)$, is the sum of all eigenvalues, \ie, $
 \text{tr}\left(\mathbf{A}\right)=\sum_{i=1}^{N}{\lambda_i}
$. And the determinant of $\mathbf{A}$, denoted $\det\left(\mathbf{A}\right)$ or $\left|\mathbf{A}\right|$, is the product of all eigenvalues, \ie, $
\det \left(\mathbf{A}\right)=\left|\mathbf{A}\right|=\prod_{i=1}^{N}{\lambda_i}
$.


\vspace{0.1cm}\noindent{\bf Spectral Radius~\cite{rota1960note}:} In mathematics, the spectral radius $\rho\left(\cdot\right)$ of a square matrix $\mathbf{A}$ is the maximum of the absolute values of its eigenvalues: \begin{equation}
\rho\left(\mathbf{A}\right)=\max\left(\left|\lambda_1\right|,\left|\lambda_2\right|,\ldots,\left|\lambda_N\right| \right).
\label{eq:spectral_radius}
\end{equation} The eigenvector corresponding to the spectral radius of $\mathbf{A}$ is commonly referred to as the principal eigenvector.

\vspace{0.1cm}\noindent{\bf Condition Number~\cite{van1969condition}:}  The condition number $\kappa\left(\cdot\right)$ of a function  quantifies the sensitivity of the function's output to small perturbations in its input. When selecting the spectral radius as the matrix norm (spectral norm), the condition number of a normal matrix $\mathbf{A}$ is:
\begin{equation}
\kappa\left(\mathbf{A}\right)=\left\Vert\mathbf{A}^{-1}\right\Vert  \left\Vert\mathbf{A}\right\Vert=\frac{\rho\left(\mathbf{A}\right)}{\rho_{\min}\left(\mathbf{A}\right)}
\label{eq:condnum}
\end{equation} where $\rho_{\min}\left(\mathbf{A}\right)=\min\left(\left|\lambda_1\right|,\left|\lambda_2\right|,\ldots,\left|\lambda_N\right|\right)$ is the minimum of the absolute values of the eigenvalues. 

\vspace{0.1cm}\noindent{\bf Spectral Entropy~\cite{shannon1948mathematical, von2018mathematical, roy2007effective, wei2024large}:} Let $\mathbf{A}$ be a positive semi-definite matrix ($\forall ~0 \leq i \leq N,~ \lambda_i \ge 0$) and the trace of $\mathbf{A}$ be positive ($\text{tr}\left(\mathbf{A}\right)>0$). Then the matrix $\mathbf{K}=\frac{\mathbf{A}}{\text{tr}\left(\mathbf{A}\right)}$ satisfies $\text{tr}\left(\mathbf{K}\right)=1$. The spectral entropy $\text{H}\left(\cdot\right)$ is:
\begin{equation}
\text{H}\left(\mathbf{A}\right)=\text{tr}\left(-\mathbf{K}\ln\mathbf{K}\right)=-\sum_{i=1}^{N}{\frac{\lambda_i}{\text{tr}\left(\mathbf{A}\right)}\ln{\frac{\lambda_i}{\text{tr}\left(\mathbf{A}\right)}}}.
\label{eq:spectral_entropy}
\end{equation}

\subsection{Extrema in Spectral Analysis}
This section will provide a detailed proof that the condition number is minimized and the spectral entropy is maximized when $s_1 = s_2 = s_3$.

\vspace{0.1cm}\noindent{\bf Condition Number: }  It is evident that the spectrum of the Gaussian satisfies the following inequality: 
\begin{equation}
0<\min\left(s_1^2, s_2^2, s_3^2\right) \le \max\left(s_1^2, s_2^2, s_3^2\right).
\end{equation} Consequently, the condition number satisfies 
\begin{equation}
\kappa\left(\boldsymbol{\Sigma}\right)=\frac{\max\left(s_1^2, s_2^2, s_3^2\right)}{\min\left(s_1^2, s_2^2, s_3^2\right)} \ge 1,
\end{equation} 
with equality iff $\max\left(s_1^2, s_2^2, s_3^2\right) = \min\left(s_1^2, s_2^2, s_3^2\right)$. Therefore, we can conclude that the condition number of $\boldsymbol{\Sigma}$ is minimized when $s_1 = s_2 = s_3$.

\vspace{0.1cm}\noindent{\bf Spectral Entropy: } Using the 3D Gaussian as an example, we aim to find the maximum value of the spectral entropy in Equation~\ref{eq:gs_entropy} of the paper. Let $t_i$ be $\frac{s_i^2}{\sum_{j=1}^{3}s_j^2}$. It is evident that this is a constrained optimization problem
\begin{gather}
\arg\max\limits_{t_1,t_2,t_3}\text{H}\left(\boldsymbol{\Sigma}\right)=-\sum_{i=1}^{3}t_i\ln{t_i}, \quad\text{ s.t.}\sum_{i=1}^{3}t_i=1
\end{gather}
 which can be solved using Lagrange multipliers as below:
 \begin{gather}
\arg\max\limits_{t_1,t_2,t_3}\mathcal{F}\left(t_1,t_2,t_3,\lambda\right)=-\sum_{i=1}^{3}t_i\ln{t_i}+\lambda\left(\sum_{i=1}^{3}t_i-1\right).
\end{gather}
Now we can calculate the gradient:
\begin{equation}
\nabla_{t_1,t_2,t_3,\lambda}\mathcal{F}\left(t_1,t_2,t_3,\lambda\right)=\left(\frac{\partial \mathcal{F}}{t_1}, \frac{\partial \mathcal{F}}{t_2},  \frac{\partial \mathcal{F}}{t_3},  \frac{\partial \mathcal{F}}{\lambda}\right)
\end{equation}
and therefore: \begin{equation}
\nabla_{t_1,t_2,t_3,\lambda}\mathcal{F}\left(t_1,t_2,t_3,\lambda\right)=\mathbf{0} \iff \begin{cases}
\lambda=\ln{t_1}+1 \\
\lambda=\ln{t_2}+1 \\
\lambda=\ln{t_3}+1 \\
\sum_{i=1}^{3}t_i=1
    \end{cases}\text{.}
\end{equation} In summary, we can conclude that the spectral entropy of $\boldsymbol{\Sigma}$ is maximized when $t_1 = t_2 = t_3$ ($s_1 = s_2 = s_3$).

\subsection{Relationship Between $\kappa\left(\cdot\right)$ and $\text{H}\left(\cdot\right)$}
\label{subsec:relation_kappa_ent}

\begin{figure}[!h]
\centering
  \subfloat [$\text{H}\left(\boldsymbol{\Sigma}\right)=f\left(\kappa\left(\boldsymbol{\Sigma}\right)\right)$ for the 2D Gaussian] {
  \label{fig:kappa_ent2d}
\includegraphics[width=1\linewidth]{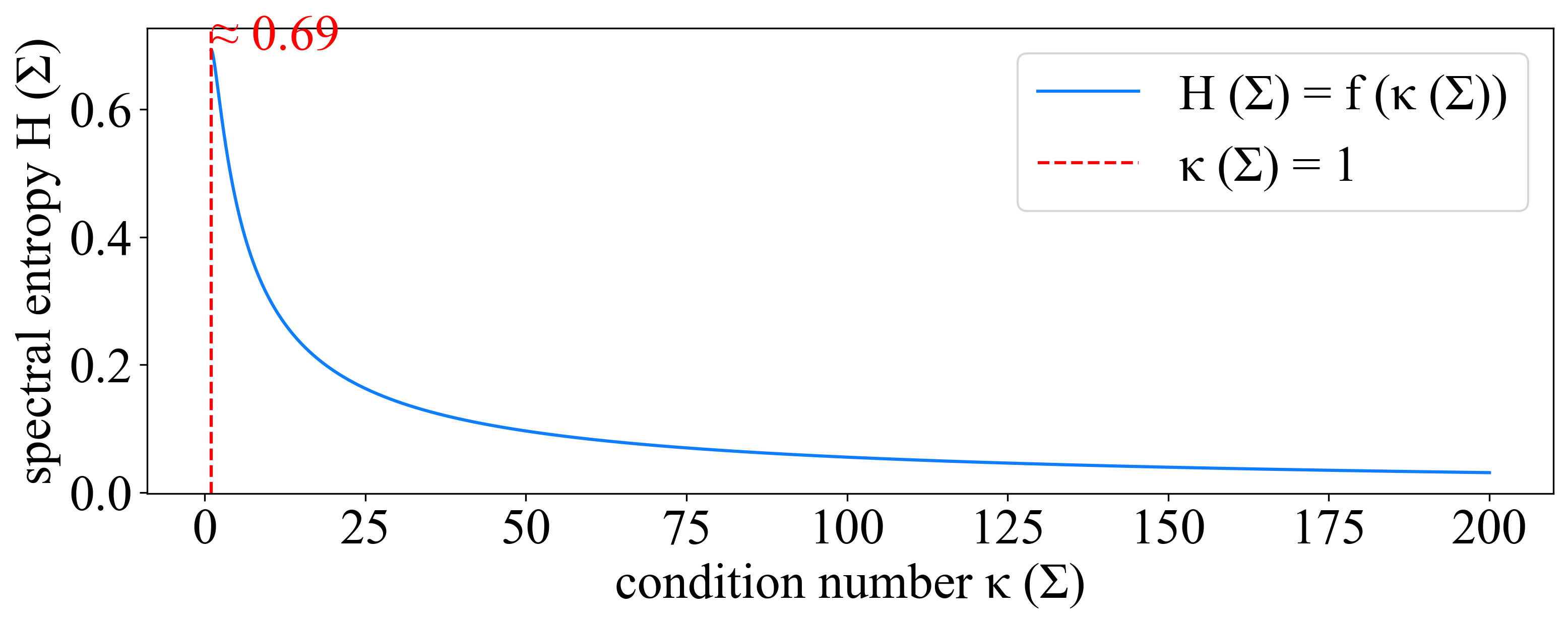}
  }\\
  \subfloat [$\text{H}\left(\boldsymbol{\Sigma}\right)=f\left(\kappa\left(\boldsymbol{\Sigma}, \lambda\right)\right)$ for the 3D Gaussian] {
  \label{fig:kappa_ent3d}
\includegraphics[width=0.7\linewidth]{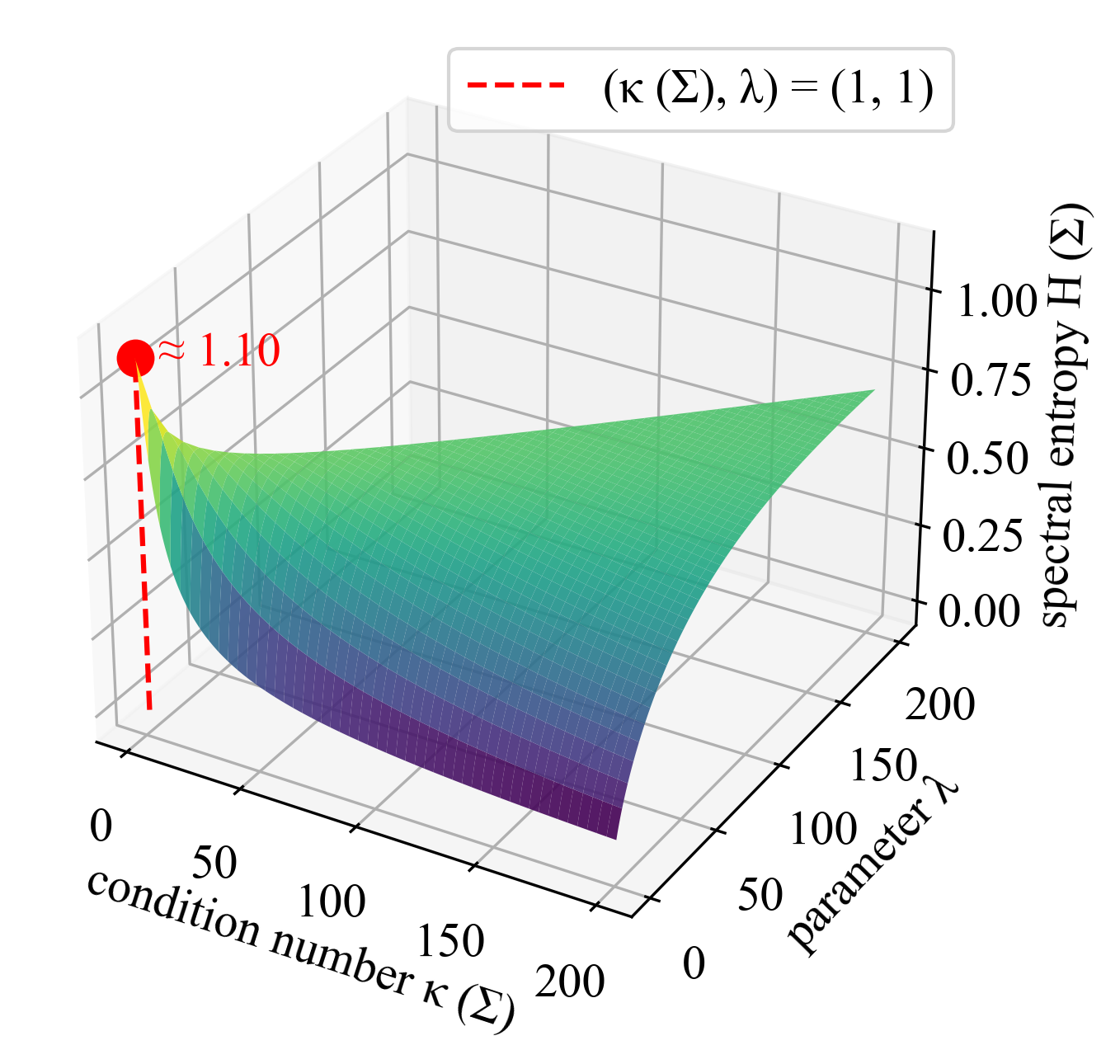}
  }

\caption{Visualization of the function $\text{H}\left(\boldsymbol{\Sigma}\right)=f\left(\kappa\left(\boldsymbol{\Sigma}\right)\right)$ for the 2D Gaussian and $\text{H}\left(\boldsymbol{\Sigma}\right)=f\left(\kappa\left(\boldsymbol{\Sigma}, \lambda\right)\right)$ for the 3D Gaussian.} 
\label{fig:kappa_ent}

\end{figure}

 We can express the spectral entropy $\text{H}\left(\boldsymbol{\Sigma}\right)$ as a function of the condition number $\kappa\left(\boldsymbol{\Sigma}\right)$ for the 2D Gaussian:
\begin{equation}
\begin{split}
\text{H}\left(\boldsymbol{\Sigma}\right)&=-\frac{s_1^2}{s_1^2+s_2^2}\ln{\frac{s_1^2}{s_1^2+s_2^2}}-\frac{s_2^2}{s_1^2+s_2^2}\ln{\frac{s_2^2}{s_1^2+s_2^2}}\\
&=\ln\left(\kappa\left(\boldsymbol{\Sigma}\right)+1\right)-\frac{\kappa\left(\boldsymbol{\Sigma}\right)\ln{\kappa\left(\boldsymbol{\Sigma}\right)}}{\kappa\left(\boldsymbol{\Sigma}\right)+1}.
\end{split}
\end{equation} Then we calculate the derivative:
\begin{equation}
\begin{split}
\frac{\mathrm{d} \text{H}\left(\boldsymbol{\Sigma}\right)}{\mathrm{d} \kappa\left(\boldsymbol{\Sigma}\right)}=-\frac{\ln{\kappa\left(\boldsymbol{\Sigma}\right)}}{\left(\kappa\left(\boldsymbol{\Sigma}\right)+1\right)^2}\le 0\text{,}~~\text{where}~\kappa\left(\boldsymbol{\Sigma}\right)\ge1,
\end{split}
\end{equation} which indicates that the spectral entropy of a 2D Gaussian decreases as the condition number increases. We visualize this function in Figure~\ref{fig:kappa_ent2d}.

For the 3D Gaussian, assuming without loss of generality that $s_1 \leq s_2 \leq s_3$, 
 we have $\kappa\left(\boldsymbol{\Sigma}\right) = \frac{s_3^2}{s_1^2}$ and let $\lambda = \frac{s_2^2}{s_1^2}$. We can similarly derive the function for the 3D Gaussian:
 \begin{equation}
\text{H}\left(\boldsymbol{\Sigma}\right)=\ln\left(\kappa\left(\boldsymbol{\Sigma}\right)+\lambda+1\right)-\frac{{\kappa\left(\boldsymbol{\Sigma}\right)}\ln{\kappa\left(\boldsymbol{\Sigma}\right)+\lambda\ln{\lambda}}}{\kappa\left(\boldsymbol{\Sigma}\right)+\lambda+1}
 \end{equation}
 and visualize the function in Figure~\ref{fig:kappa_ent3d}.

\subsection{View-Inconsistency in Filtering}
\label{subsec:append_view_incons}

The Jacobian matrix of the local affine approximation is as follows:

\begin{equation}
\begin{split}
\mathbf{J}&=\begin{bmatrix}
\frac{f_x}{\mu_z} & 0 & -\frac{f_x\mu_x}{\mu_z^2} \\
0 & \frac{f_y}{\mu_z} & -\frac{f_y\mu_y}{\mu_z^2} \\
\end{bmatrix}
=\left[\begin{matrix}
\frac{f_x}{\mu_z} & 0 & 0 \\
0 & \frac{f_y}{\mu_z} & 0 \\
\end{matrix}\right]\left[\begin{matrix}
1 & 0 & -\frac{\mu_x}{\mu_z} \\
0 & 1 & -\frac{\mu_y}{\mu_z} \\
0 & 0 & 1
\end{matrix}\right]
\end{split}
\end{equation} where $f_x$, $f_y$ denote the intrinsic parameters of the camera model and $\boldsymbol{\mu}^{'}=\left[\begin{matrix}\mu_x & \mu_y & \mu_z\end{matrix}\right]^{\top}$ is the position of the 3D Gaussian in the camera space. We assume that the position of the Gaussian projected onto the $z=1$ plane, \ie,  $\left[\frac{\mu_x}{\mu_z},\frac{\mu_y}{\mu_z}\right]^{\top}$, remains unchanged during the camera zoom-in. Since the covariance matrix in the camera space $\boldsymbol{\Sigma}^{'}$ also remains unchanged when zooming in, the following matrix is a constant matrix 
\begin{equation}
\begin{split}
    \boldsymbol{\Sigma}^{''}=\left[\begin{matrix}
1 & 0 & -\frac{\mu_x}{\mu_z} \\
0 & 1 & -\frac{\mu_y}{\mu_z} \\
0 & 0 & 1
\end{matrix}\right]    \boldsymbol{\Sigma}^{'}\left[\begin{matrix}
1 & 0 & -\frac{\mu_x}{\mu_z} \\
0 & 1 & -\frac{\mu_y}{\mu_z} \\
0 & 0 & 1
\end{matrix}\right]^{\top}=\begin{bmatrix}
        a & b & c\\
        b & d & e\\
        c & e & f
    \end{bmatrix}
\end{split}
\end{equation} where $a,b,c,d,e,f\in\left[0,+\infty\right)$. Then we obtain:
\begin{gather}
\boldsymbol{\Sigma}_{\text{filter}}=\mathbf{J}\boldsymbol{\Sigma}^{'}\mathbf{J}^{\top}+s\mathbf{I}\approx\left[\begin{matrix}s + \frac{a f_{x}^{2}}{\mu_{z}^{2}} & \frac{b f_{x} f_{y}}{\mu_{z}^{2}}\\\frac{b f_{x} f_{y}}{\mu_{z}^{2}} & s + \frac{d f_{y}^{2}}{\mu_{z}^{2}}\end{matrix}\right]\text{.}
\end{gather}
Then we can compute the condition number of the matrix:
\begin{equation}
\begin{split}
&\kappa\left(\boldsymbol{\Sigma}_{\text{filter}}\right)=\frac{\frac{1}{2}\text{tr}\left(\boldsymbol{\Sigma}_{\text{filter}}\right)+\sqrt{\frac{1}{4}\text{tr}^2\left(\boldsymbol{\Sigma}_{\text{filter}}\right)- \left|\boldsymbol{\Sigma}_{\text{filter}}\right|}}
{\frac{1}{2}\text{tr}\left(\boldsymbol{\Sigma}_{\text{filter}}\right)-\sqrt{\frac{1}{4}\text{tr}^2\left(\boldsymbol{\Sigma}_{\text{filter}}\right)- \left|\boldsymbol{\Sigma}_{\text{filter}}\right|}}\\
&=\frac{2  s + \left(a  + d \frac{f_y^2}{f_x^2} + \sqrt{\left(a  - d \frac{f_y^2}{f_x^2}\right)^2 + 4 b^{2}  \frac{f_y^2}{f_x^2} }\right)\frac{f_x^2}{\mu_z^2}}{2  s + \left(a  + d \frac{f_y^2}{f_x^2} - \sqrt{\left(a  - d \frac{f_y^2}{f_x^2}\right)^2 + 4 b^{2}  \frac{f_y^2}{f_x^2} }\right)\frac{f_x^2}{\mu_z^2}}
\end{split}
\end{equation} which is a function of $\frac{f_x^2}{\mu_z^2}$ due to the other values remaining constant during the camera zoom-in. And the derivative of the function is:
\begin{equation}
\frac{\mathrm{d}\kappa\left(\boldsymbol{\Sigma}_{\text{filter}}\right)}{\mathrm{d}\frac{f_x^2}{\mu_z^2}}=
\frac{4{p}s}
{\left(2  s + \left(a  + d \frac{f_y^2}{f_x^2} - p\right)\frac{f_x^2}{\mu_z^2}\right)^2}>0
\end{equation} where $p=\sqrt{\left(a  - d \frac{f_y^2}{f_x^2}\right)^2 + 4 b^{2}  \frac{f_y^2}{f_x^2} }$. This indicates that the condition number of a 2D Gaussian
increases when zooming in the camera (increasing the $\frac{f_x}{\mu_z}$). And according to 
 the relationship between spectral entropy and the condition number derived in Section~\ref{subsec:relation_kappa_ent}, the spectral entropy $\text{H}\left(\boldsymbol{\Sigma}_{\text{filter}}\right)$ also decreases when zooming in the camera. The curve of $\kappa\left(\boldsymbol{\Sigma}_{\text{filter}}\right)$  as a function of $\frac{f_x^2}{\mu_z^2}$ is visualized in Figure~\ref{fig:zoomin} of the paper.

\subsection{$k$ in 3D Shape-Aware Splitting}

For the 3D Gaussian, assuming without loss of generality
that $s_1 \leq s_2 \leq s_3$, we can derive the following:
\begin{gather}
    \kappa\left(\boldsymbol{\Sigma}_{\text{split}}\right)=
    \frac{\max\left(\frac{s_3^2}{\left(k+k_0\right)^2}, \frac{s_2^2}{k_0^2}\right)}{\min\left(\frac{s_3^2}{\left(k+k_0\right)^2}, \frac{s_1^2}{k_0^2}\right)}=\begin{cases}
    \frac{k_0^2 s_3^2}{\left(k+k_0\right)^2 s_1^2}\\
    \frac{s_2^2}{s_1^2}\\
    \frac{\left(k+k_0\right)^2 s_2^2}{k_0^2 s_3^2}\\
    \end{cases},\\
    \kappa\left(\boldsymbol{\Sigma}\right)=\frac{s_3^2}{s_1^2}.
\end{gather}
Clearly, we have:
\begin{equation}
    \frac{k_0^2 s_3^2}{\left(k+k_0\right)^2 s_1^2}<\frac{s_3^2}{s_1^2}=\kappa\left(\boldsymbol{\Sigma}\right),\quad \frac{s_2^2}{s_1^2}\leq\frac{s_3^2}{s_1^2}=\kappa\left(\boldsymbol{\Sigma}\right).
\end{equation}
And when Equation~\ref{eq:k_condition} in the paper is satisfied, we can derive the following: 
\begin{gather}
    \frac{k+k_0}{k_0}<\frac{\rho^{\frac{3}{2}}\left(\boldsymbol{\Sigma}\right)}{\sqrt{\left|\boldsymbol{\Sigma}\right|}}=\frac{s_3^2}{ s_2 s_1},\\
\frac{\left(k+k_0\right)^2 s_2^2}{k_0^2 s_3^2}<\frac{\left(\frac{s_3^2}{s_2 s_1}\right)^2 s_2^2}{ s_3^2}=\frac{s_3^2}{s_1^2}=\kappa\left(\boldsymbol{\Sigma}\right). 
\end{gather} In summary, we can conclude that the condition number after splitting does not exceed that before splitting, \ie, $\kappa\left(\boldsymbol{\Sigma}_{\text{split}}\right)\leq\kappa\left(\boldsymbol{\Sigma}\right)$, when Equation~\ref{eq:k_condition} in the paper is satisfied.

\subsection{2D View-Consistent Filtering Kernel}

We prove the view-consistency of our filter:
\begin{equation}
 \begin{split}
\kappa\left(\boldsymbol{\Sigma}_{\text{train}}\right)&=\frac{\rho\left(\mathbf{J}_{\text{train}}\boldsymbol{\Sigma}^{'}\mathbf{J}_{\text{train}}^{\top}\right)+\rho\left(s\mathbf{I}\right)}{\rho_{\min}\left(\mathbf{J}_{\text{train}}\boldsymbol{\Sigma}^{'}\mathbf{J}_{\text{train}}^{\top}\right)+\rho_{\min}\left(s\mathbf{I}\right)}\\
&=\frac{\rho\left(\boldsymbol{\Sigma}^{'}\right)+\rho\left(\mathbf{J}_{\text{train}}^{-1}s\mathbf{I}\left(\mathbf{J}_{\text{train}}^{-1}\right)^{\top}\right)}{\rho_{\min}\left(\boldsymbol{\Sigma}^{'}\right)+\rho_{\min}\left(\mathbf{J}_{\text{train}}^{-1}s\mathbf{I}\left(\mathbf{J}_{\text{train}}^{-1}\right)^{\top}\right)}\\
&=\frac{\rho\left(\boldsymbol{\Sigma}_{\text{test}}\right)+\rho\left(\left(\mathbf{J}_{\text{test}}\mathbf{J}_{\text{train}}^{-1}\right)s\mathbf{I}\left(\mathbf{J}_{\text{test}}\mathbf{J}_{\text{train}}^{-1}\right)^{\top}\right)}{\rho_{\min}\left(\boldsymbol{\Sigma}_{\text{test}}\right)+\rho_{\min}\left(\left(\mathbf{J}_{\text{test}}\mathbf{J}_{\text{train}}^{-1}\right)s\mathbf{I}\left(\mathbf{J}_{\text{test}}\mathbf{J}_{\text{train}}^{-1}\right)^{\top}\right)}\\
&=\kappa\left(\boldsymbol{\Sigma}_{\text{test}}\right).
 \end{split} 
\end{equation}
In Section~\ref{subsec:append_view_incons}, when $s\propto \frac{f_x^2}{\mu_z^2} $, the function $\kappa\left(\boldsymbol{\Sigma}_{\text{filter}}\right)$ 
 and $\text{H}\left(\boldsymbol{\Sigma}_{\text{filter}}\right)$ are constant. Therefore, we can approximate this operation using the filter kernel function $s\left(f_x, \mu_z\right)=s_0\frac{f_x^2}{\mu_z^2}$, where $s_0$ is a constant.
 
\section{Spectral-GS Algorithm}
Our 3D shape-aware splitting and 2D view-consistent filtering algorithms are summarized in Algorithm \ref{alg:spectral_gs}.
	\begin{algorithm}[!h]
		\caption{Spectral-GS Algorithm\\
		$W$, $H$: width and height of the training or testing images}
		\label{alg:spectral_gs}
		\begin{algorithmic}
			\State $M, S, C, O \gets$ Gaussians() \Comment{Pos, Covs, Colors, Opacs}

            \If{is testing}
            \State $V \gets$ TestingView()	 
               \State $M_p$, $S_p$ $\gets$ SplatGaussian($W$, $H$, $M$, $S$, $V$)
                \State $S_f$, $O_f$ $\gets$ SpectralBasedFilter($S_p$, $O$, $V$) \Comment{Filter}
			\State $I \gets$ Rasterize($W$, $H$, $M_p$, $S_f$, $C$, $O_f$, $V$)	
            \Else
			\State $i \gets 0$	\Comment{Iteration Count}
			\While{not converged}
			\State $V, \hat{I} \gets$ TrainingView()	 \Comment{Camera and Image }
                \State $M_p$, $S_p$ $\gets$ SplatGaussian($W$, $H$, $M$, $S$, $V$)
                \State $S_f$, $O_f$ $\gets$ SpectralBasedFilter($S_p$, $O$, $V$) \Comment{Filter}
			\State $I \gets$ Rasterize($W$, $H$, $M_p$, $S_f$, $C$, $O_f$, $V$)	
			
			\State $\mathcal{L} \gets$ Loss($I, \hat{I}$) \Comment{Compute Loss}
			
			\State $M$, $S$, $C$, $O$ $\gets$ Adam($\nabla \mathcal{L}$)  \Comment{Backprop \& Step}

			\If{IsRefinementIteration($i$)}
			\ForAll{ $\mathcal{G}^{3D}(\boldsymbol{\mu}, \boldsymbol{\Sigma}, c, o)$ $\textbf{in}$ $(M, S, C, O)$}
			\If{$o < \epsilon_{\text{o}}$ or IsInvalidSpectrum($o$, $\boldsymbol{\Sigma}$) }	
			\State SpectralBasedPruneGaussian()
			\EndIf
			\If{$\Vert\nabla_{\boldsymbol{\mu}_{\text{proj}}} \mathcal{L}\Vert_F > \tau_{\text{loss}}$} \Comment{Densify}
			\If{$\rho\left(\boldsymbol{\Sigma}\right)>\tau_{\text{radius}}$}	\Comment{Split}
			\State LossBasedSplitGaussian($\boldsymbol{\mu}, \boldsymbol{\Sigma}, c, o$)
			\Else								\Comment{Clone}
			\State LossBasedCloneGaussian($\boldsymbol{\mu}, \boldsymbol{\Sigma}, c, o$)
			\EndIf	
			\EndIf
            \If{$\text{H}\left(\boldsymbol{\Sigma}\right) < \tau_{\text{spectral}}$} \Comment{Densify}
            \State    SpectralBasedSplitGaussian($\boldsymbol{\mu}, \boldsymbol{\Sigma}, c, o$)
            \EndIf
			\EndFor		
			\EndIf
			\State $i \gets i+1$
			\EndWhile
            \EndIf
		\end{algorithmic}
	\end{algorithm}


\section{Additional Results}

\begin{figure*}[!ht]
\centering
\includegraphics[width=1\textwidth]{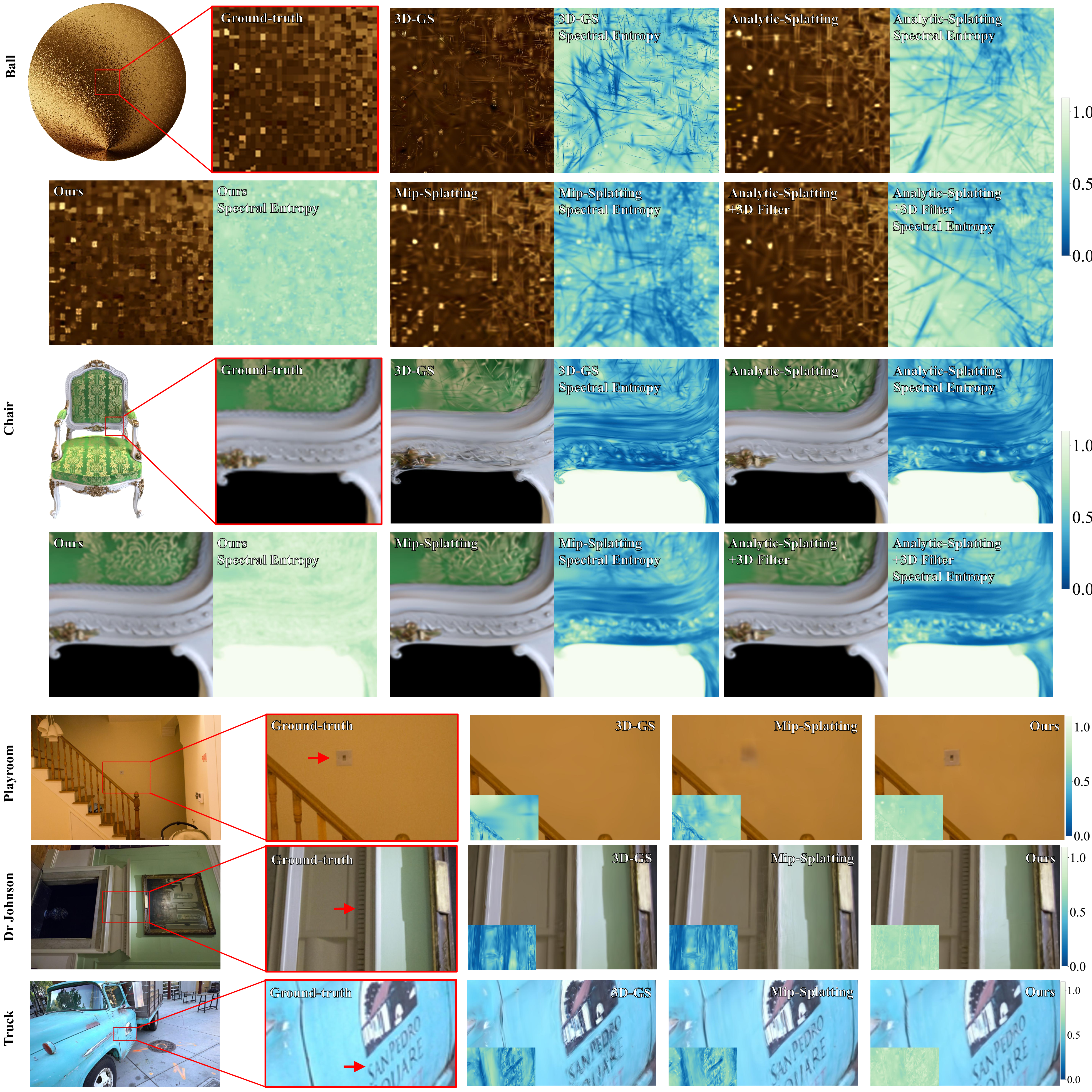}
\caption{We show comparisons of our method to previous methods and the corresponding ground truth images from held-out test views. Additionally, We visualize the spectral entropy maps of 3D Gaussians after optimization. Bluer regions indicate lower spectral entropy, with more needle-like degraded Gaussians, while greener regions represent higher spectral entropy, without noticeable needle-like artifacts.} 
\label{fig:sup_cmp}

\end{figure*}

\begin{figure*}[!ht]
\centering
\includegraphics[width=0.85\textwidth]{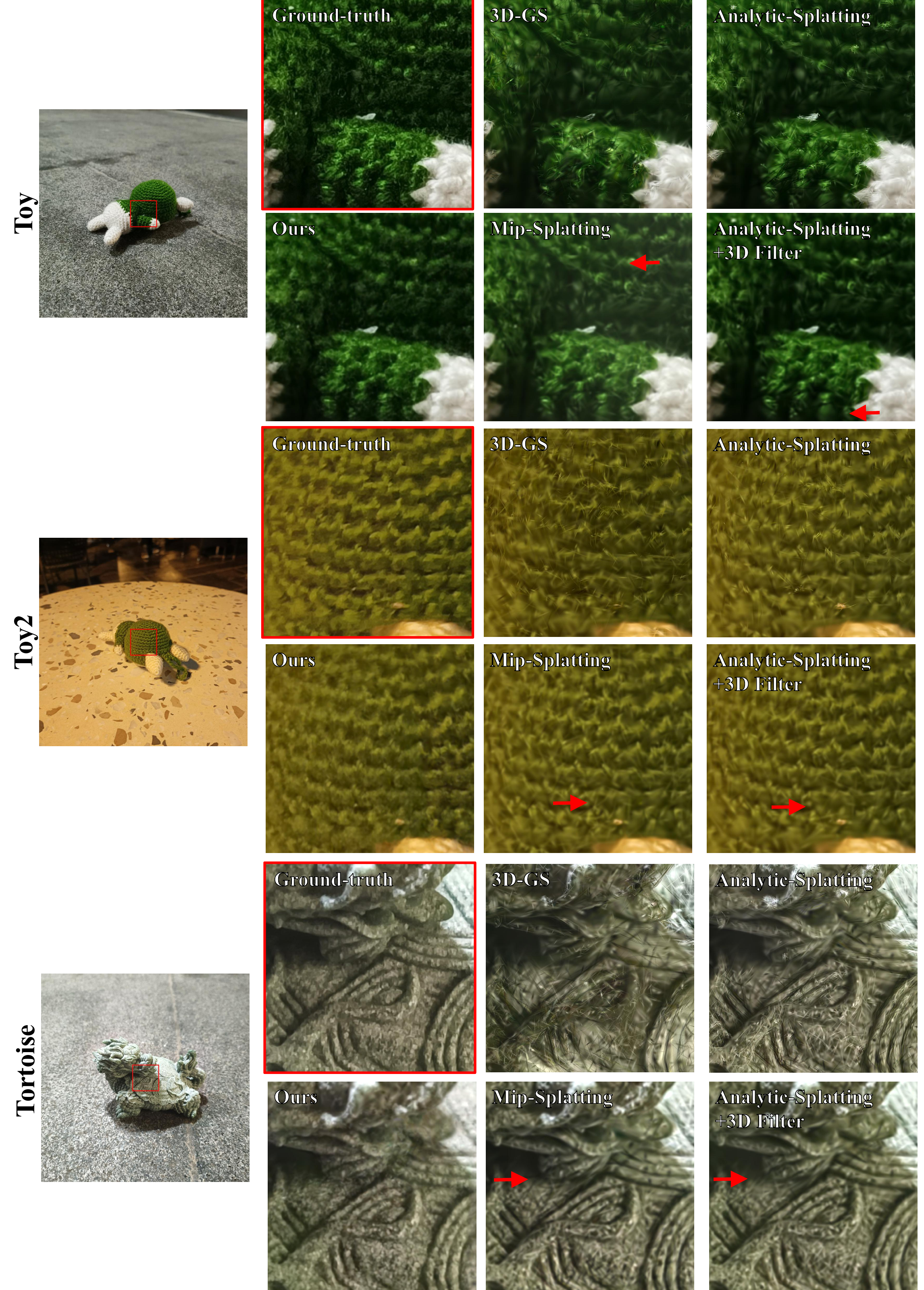}
\caption{We show comparisons of our method to previous methods and the corresponding ground truth images from held-out test views.  Differences in quality highlighted
by arrows/insets.} 
\label{fig:sup_cmp2}

\end{figure*}

Figures~\ref{fig:sup_cmp} and ~\ref{fig:sup_cmp2} illustrate additional qualitative comparisons. The scenes in Figure~\ref{fig:sup_cmp} are \textsc{Ball} with modified textures, \textsc{Chair}, \textsc{Playroom} and \textsc{DrJohnson} from the \textit{Deep Blending}~\cite{hedman2018deep}, and \textsc{Truck}. The scenes in Figure~\ref{fig:sup_cmp2} are from the \textit{High-frequency Spectrum} dataset captured by us. 3D-GS~\cite{kerbl20233d} and Analytic Splatting~\cite{liang2024analytic} exhibit pronounced needle-like artifacts. In contrast, methods with the 3D smoothing filter~\cite{mip_gs} show reduced needle-like Gaussian artifacts due to the 3D filter increasing the spectral entropy of the Gaussians. Our Spectral-GS eliminates the needle-like artifacts and enhances details, achieving more photorealistic rendering.

\end{document}